\documentclass{article}

\usepackage[nonatbib,preprint]{neurips_2021}

\usepackage[utf8]{inputenc} % allow utf-8 input
\usepackage[T1]{fontenc}    % use 8-bit T1 fonts
\usepackage{url}            % simple URL typesetting
\usepackage{booktabs}       % professional-quality tables
\usepackage{amsfonts}       % blackboard math symbols
\usepackage{nicefrac}       % compact symbols for 1/2, etc.
\usepackage{microtype}      % microtypography
\usepackage{xcolor}         % colors
\usepackage{epsfig}
\usepackage{graphicx}
\usepackage{subfigure}
\usepackage{multirow}
\usepackage{wrapfig}
\usepackage{amsbsy}
\usepackage{arydshln}
\usepackage{algorithm}
\usepackage{algorithmic}
\usepackage{amsmath}

\usepackage{hyperref}   
\usepackage{pifont}% http://ctan.org/pkg/pifont
\newcommand{\cmark}{\ding{51}}%
\newcommand{\xmark}{\ding{55}}%

\newtheorem{prop}{Proposition}

\newcommand{\R}{\mathbb{R}}
\newcommand{\LL}{\mathcal{L}}
\newcommand{\dimt}{\R^{N\times C\times H\times W}}
\newcommand{\dimk}{\R^{C\times K\times K\times C}}
\newcommand{\Sigmainv}{\mathbf{\Sigma}^{-\frac{1}{2}}}
\newcommand{\SigmainvN}{\mathbf{\Sigma}_{N}^{-\frac{1}{2}}}

\newcommand{\tran}{{^{\mkern-1.5mu\mathsf{T}}}}
\newcommand{\tranT}{{^{\mathsf{T}}}}

\newcommand{\ie}{\textit{i.e.}$\,$}
\newcommand{\eg}{\textit{e.g.}}

\title{BWCP: Probabilistic Learning-to-Prune Channels for ConvNets via Batch Whitening}

\author{%
  Wenqi Shao \\
  The Chinese University of Hong Kong\\
  \texttt{weqish@link.cuhk.edu.hk} \\
   \And
   Hang Yu \\
   Huawei Noah’s Ark Lab\\
  \texttt{hyu0829@buaa.edu.cn} \\
  \And
   Zhaoyang Zhang \\
  The Chinese University of Hong Kong\\
  \texttt{zhaoyangzhang@link.cuhk.edu.hk} \\
  \And
   Hang Xu \\
   Huawei Noah’s Ark Lab\\
  \texttt{chromexbjxh@gmail.com} \\%
  \And
   Zhenguo Li \\
   Huawei Noah’s Ark Lab\\
  \texttt{ li.zhenguo@huawei.com} \\
  \And
   Ping Luo \\
  The University of Hong Kong\\
  \texttt{pluo@cs.hku.hk} \\
}

\begin{document}

\maketitle

\begin{abstract}
  This work presents a probabilistic channel pruning method to accelerate Convolutional Neural Networks (CNNs). Previous pruning methods often zero out unimportant channels in training in a deterministic manner, which reduces CNN's learning capacity and results in suboptimal performance. To address this problem, we develop a probability-based pruning algorithm, called batch whitening channel pruning (BWCP), which can stochastically discard unimportant channels by modeling the probability of a channel being activated. BWCP has several merits. (1) It simultaneously trains and prunes CNNs from scratch in a probabilistic way, exploring larger network space than deterministic methods. (2) BWCP is empowered by the proposed batch whitening tool, which is able to empirically and theoretically increase the activation probability of useful channels while keeping unimportant channels unchanged without adding any extra parameters and computational cost in inference. (3) Extensive experiments on CIFAR-10, CIFAR-100, and ImageNet with various network architectures show that BWCP  outperforms its counterparts by achieving better accuracy given limited computational budgets. For example, ResNet50 pruned by BWCP has only 0.70\% Top-1 accuracy drop on ImageNet, while reducing 43.1\% FLOPs of the plain ResNet50.

\end{abstract}

\section{Introduction}
Deep convolutional neural networks (CNNs) have achieved superior performance in a variety of computer vision tasks such as image recognition~\cite{he2016deep, he2016identity}, object detection~\cite{ren2017faster, he2020mask}, and semantic segmentation~\cite{chen2018deeplab,shelhamer2017fully}.
However, despite their great success, deep CNN models often have massive demand on storage, memory bandwidth, and computational power \cite{han2018bandwidth}, making them difficult to be plugged onto resource-limited platforms, such as portable and mobile devices \cite{deng2020model, han2018bandwidth}.
Therefore, proposing efficient and effective model compression methods has become a hot research topic in the deep learning community \cite{zhu2018to,  sun2017ensemble}.

Model pruning, as one of the vital model compression techniques,  has been extensively investigated. It reduces model size and computational cost by removing unnecessary or unimportant weights or channels in a CNN~\cite{han2016deep, wen2016learning, han2015learning, li2016pruning}.
For example, many recent works \cite{han2016deep, han2015learning, guo2016dynamic} prune fine-grained weights of filters. \cite{han2015learning} proposes to discard the weights that have magnitude less than a predefined threshold. \cite{guo2016dynamic} further utilizes a sparse mask on a weight basis to achieve pruning. Although these unstructured pruning methods attempt to solve the problems such as optimal pruning schedule, 
%and better heuristics to represent the importance of weights in the model, 
they do not take the structure of CNNs into account, 
%thus failing to help in the scaling of 
preventing them from being accelerated on hardware such as GPU for parallel computations \cite{liu2018rethinking,wen2016learning}.
\begin{figure}
\begin{center}
\includegraphics[scale=0.45]{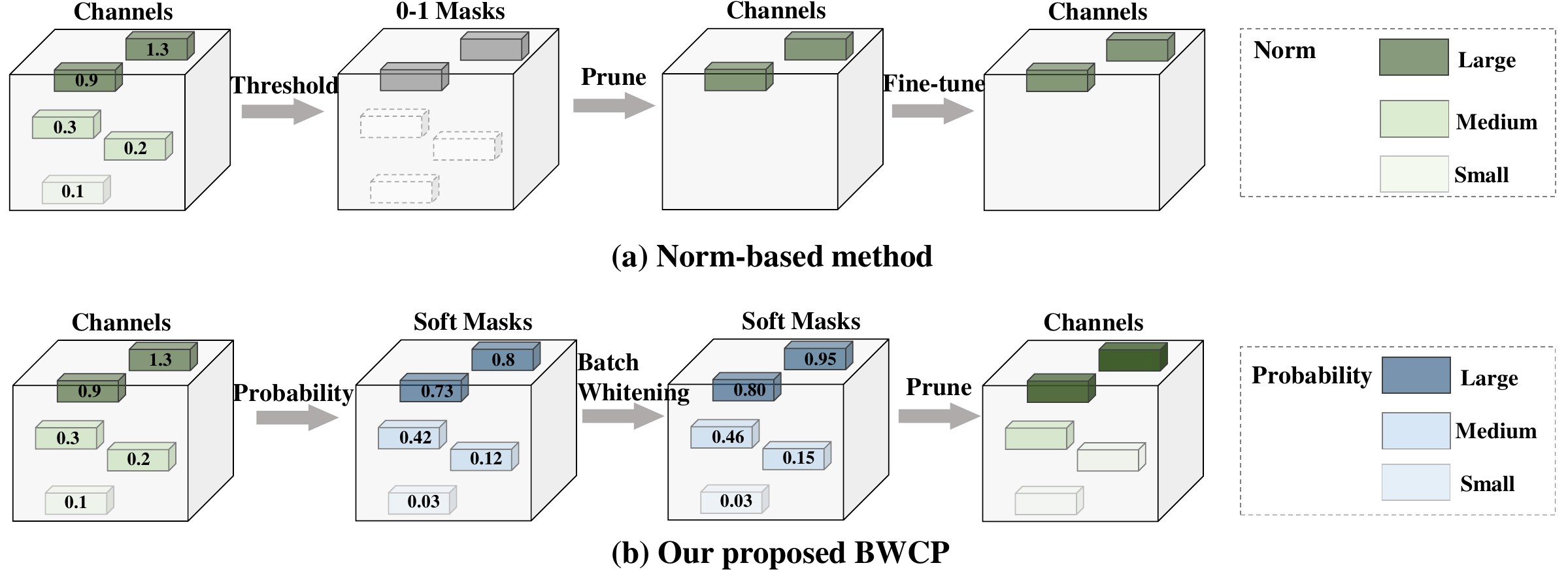}
\end{center}
\vspace{-0.1in}
  \caption{Illustration of our proposed BWCP. (\textbf{a}) Previous channel pruning methods utilize a unified criterion such as the norm ~\cite{liu2017learning} and rank \cite{lin2020hrank} of channels to deterministically remove unimportant channels, which deteriorates performance and needs a extra fine-tuning process\cite{frankle2018lottery}.
  %Such a ``smaller-norm-less-important" criterion requires the norm deviation between channels to be large enough so that less critical channels can be easily selected ~\cite{he2019filter}.
  (\textbf{b}) Our proposed BWCP is a probability-based pruning framework where unimportant channels are stochastically pruned with activation probability, thus maintaining the learning capacity of original CNNs and achieving good performance without fine-tuning.
  In particular, our proposed batch whitening (BW) tool can increase the activation probability of useful channels while keeping the activation probability of unimportant channels unchanged, enabling BWCP to identify unimportant channels reliably.}
\vspace{-0.2in}
\label{fig:Fig1}
\end{figure}
To achieve efficient model storage and computations, we focus  on structured channel pruning \cite{wen2016learning, yang2019deephoyer, liu2017learning}, which removes entire structures in a CNN such as filter or channel. %\footnote{Since a filter in the convolution kernel corresponds to an output channel, we do not distinguish filter pruning from channel pruning.}. 
A typical structured channel pruning approach commonly contains three stages, including pre-training a full model, pruning unimportant channels by the predefined criteria such as $\ell_p$ norm, and fine-tuning the pruned model \cite{liu2017learning, luo2017thinet}, as shown in Fig.\ref{fig:Fig1} (a). However, it is usually hard to find a global pruning threshold to select unimportant channels, because the norm deviation between channels is often too small \cite{he2019filter}. More importantly, as some channels are permanently zeroed out in the pruning stage, such a multi-stage procedure usually not only relies on hand-crafted heuristics but also limits the learning capacity \cite{he2018soft, he2019filter}.

To tackle the above issues, we propose a simple but effective probability-based channel pruning framework, named batch-whitening channel pruning (BWCP), where unimportant channels are pruned in a stochastic manner, thus preserving the channel space of CNNs in training (\ie~ the diversity of CNN architectures is preserved). To be specific, as shown in Fig.\ref{fig:Fig1} (b), we assign each channel with an activation probability (\ie the probability of a channel being activated), by exploring the properties of the batch normalization layer \cite{ioffe2015batch, arpit2016normalization}. A 
larger activation probability indicates that the corresponding channel is more likely to be preserved.
% with a high probability.
%

We also introduce a capable tool, termed batch whitening (BW), which can increase the activation probability of useful channels, while keeping the unnecessary channels unchanged. By doing so, the deviation of the activation probability  between channels is explicitly enlarged, enabling   BWCP to identify unimportant channels during training easily.
Such an appealing property is justified by theoretical analysis and experiments. 
Furthermore,
we exploit activation probability adjusted by BW to generate a set of differentiable masks by a soft sampling procedure with Gumbel-Softmax technique, allowing us to train BWCP in an online ``pruning-from-scratch'' fashion stably. After training, we obtain the final compact model by directly discarding the channels with zero masks.

The main \textbf{contributions} of this work are three-fold. (1) We propose a probability-based channel pruning framework BWCP, which explores a larger network space than deterministic methods. (2) BWCP can easily identify unimportant channels by adjusting their activation probabilities without adding any extra model parameters and computational cost in inference. (3) Extensive experiments on CIFAR-10, CIFAR-100 and ImageNet datasets with various network architectures show that BWCP can achieve better recognition performance given the comparable amount of resources compared to existing approaches. For example, BWCP can reduce 68.08\% Flops by compressing 93.12\% parameters of VGG-16 with merely accuracy drop and ResNet-50 pruned by BWCP has only 0.70\% top-1 accuracy drop on ImageNet while reducing 43.1\% FLOPs.

\vspace{-0.1in}
\section{Related Work}
%
%There is a tremendous surge of research activity in compressing deep neural networks. 
%In this section, we introduce some prevailing pruning techniques. %Particularly, we analyze the similarities and differences between the proposed BWCP algorithm and previous pruning methods.

%\textbf{Model Compression.} By now, there are a large number of model compression approaches such as low-rank factorization \cite{lin2018holistic,yu2017compressing}, network quantization \cite{polino2018model,han2016deep} and weight or channel pruning \cite{wen2016learning, yang2019deephoyer, liu2017learning}. Typically, low-rank factorization reduces float point operations (FLOPs) by decomposing the original weight matrix into several submatrices with low rank. For example, Singular Value Decomposition (SVD)\cite{denton2014exploiting} and Tucker Decomposition \cite{kim2015compression} are utilized to compress the weights of convolution filters. Another line of research in compressing deep model comes in the form of model quantization \cite{han2016deep,polino2018model}, which represents network parameters by a low-bit precision, thus reducing the computational cost and model size. Different from these methods, this work focuses on pruning techniques.

\textbf{Weight Pruning.}
Early network pruning methods mainly remove the unimportant weights in the network. For instance,  Optimal Brain Damage \cite{lecun1990optimal} measures the importance of weights by evaluating the impact of weight on the loss function and prunes less important ones. However,   it is not applicable in modern network structure due to the heavy computation of the Hessian matrix. Recent work assesses the importance of the weights through the magnitude of the weights itself. 
\cite{han2016deep} iteratively removes the weights with the magnitude less than a predefined threshold. Moreover, \cite{guo2016dynamic} prune the network by encouraging weights to become exactly zero. The computation involves weights with zero can be discarded. However, a major drawback of weight pruning techniques is that they do not take the structure of CNNs into account, thus failing to help scale pruned models on commodity hardware such as GPUs \cite{liu2018rethinking,wen2016learning}.

\textbf{Channel Pruning.}
%Unlike the above methods that prune the fine-grained weights, c
Channel pruning approaches directly prune feature maps or filters of CNNs, making it easy for hardware-friendly implementation. For instance, multiple bodies of work impose channel-level sparsity, and filters with small value are selected to be pruned. They include work on  relaxed $\ell_0$ regularization \cite{louizos2017learning} and group regularizer \cite{yang2019deephoyer}.  Some recent work also propose to  rank the importance of  filters by different criteria including $\ell_1$ norm \cite{liu2017learning, li2017}, $\ell_2$ norm \cite{frankle2018lottery} and High Rank channels \cite{lin2020hrank}. For example, 
\cite{liu2017learning} explores the importance of filters through scale parameter $\gamma$ in batch normalization. Although these approaches introduce minimum overhead to the training process, they are not trained in an end-to-end manner and usually either apply on a pre-trained model or require an extra fine-tuning procedure. 

Recent works tackle this issue by pruning CNNs from scratch. For example, FPGM \cite{he2019filter} zeros in unimportant channels and continues training them after each training epoch. Moreover, both SSS and DSA learn a differentiable binary mask which is generated by the importance of channels and require no extra fine-tuning. Our proposed BWCP is most related to variational pruning \cite{zhao2019variational} and SCP \cite{kang2020operation} as they also employ the property of normalization layer and associate the importance of channel with probability. The main difference is that our method adopts the idea of whitening to perform channel pruning. We will show that the proposed batch whitening (BW) technique can increase the activation probability of useful channels while keeping the activation probability of unimportant channels unchanged, making it easy for identifying unimportant channels.
%The main difference is that our method stochastically discards unimportant channels with probabilities generated by the proposed batch whitening (BW) technique. We will show that BW can increase the activation probability of useful channels while keeping the activation probability of unimportant channels unchanged, making it easy for identifying unimportant channels.

%

%they require that the norm deviation between channels is large~\cite{he2019filter} enough so that an appropriate threshold can found to select to-be-pruned channels. 

%\textbf{Normalization.} Normalization method deals with parameter training of deep models. There are many practices on normalizer development, such as batch normalization (BN) \cite{ioffe2015batch} and group normalization (GN) \cite{wu2018group}. A normalization scheme is typically applied after a convolution layer and contains two stages: standardization and rescaling. Previous work \cite{mehta2019implicit,frankle2020training} has revealed that parameters in BN have a close connection with channel-level sparsity.  In this work, we exploit the property that BN makes channel features normally distributed \cite{arpit2016normalization} to achieve channel pruning. 

%
\section{Preliminary}
\textbf{Notation.} We use regular letters, bold letters, and capital letters to denote scalars such
as ‘$x$’,  and  vectors (\eg vector,
matrix, and tensor) such as ‘$\mathbf{x}$’ and random variables such as  `$X$', respectively.

We begin with introducing a  building layer in recent deep neural nets which typically consists of a convolution layer, a batch normalization (BN) layer, and a rectified linear unit (ReLU) \cite{ioffe2015batch, he2016deep}. Formally, it can be written by
\begin{equation}\label{eq:conv-norm-relu}
\mathbf{x}_c = \mathbf{w}_c * \mathbf{z},\quad \tilde{\mathbf{x}}_c = \gamma_c \bar{\mathbf{x}}_c +\beta_c,\quad \mathbf{y}_c = \max\{\mathbf{0},\tilde{\mathbf{x}}_c\}
\end{equation}
%$\mathbf{w}_c\in\mathbb{R}^{C\times K \times K}$
where $c \in [C]$ denotes channel index and $C$ is channel size. In Eqn.(\ref{eq:conv-norm-relu}), `$*$' indicates convolution operation and $\mathbf{w}_c$ is filter weight corresponding to the $c$-th  output channel, \ie $\mathbf{x}_c\in\mathbb{R}^{N\times H  \times W}$. To perform normalization, $\mathbf{x}_c$ is firstly standardized to $\bar{\mathbf{x}}_c$ through $\bar{\mathbf{x}}_c=(\mathbf{x}_c-\mathbb{E}[\mathbf{x}_c])/\sqrt{\mathbb{D}[\mathbf{x}_c]}$ where $\mathbb{E}[\cdot]$ and $\mathbb{D}[\cdot]$ indicate calculating mean and variance over a batch of samples, and then is re-scaled to $\tilde{\mathbf{x}}_c$ by scale parameter $\gamma_c$ and bias $\beta_c$. Moreover, the output feature $\mathbf{y}_c$ is obtained by ReLU activation that discards the negative part of $\tilde{\mathbf{x}}_c$.

\textbf{Criterion-based channel pruning.} For channel pruning, previous methods usually employ a `small-norm-less-important' criterion to measure the importance of channels. For example, 
%\cite{he2019filter} calculate and the geometric mean of filter $\mathbf{w}_c$, and assign zero values to those filters with the small norm at each training step. It is noteworthy that 
BN layer can be applied in channel pruning \cite{liu2017learning}, where a channel with a small value of $\gamma_c$ would be removed. The reason is that the $c$-th output channel $\tilde{\mathbf{x}}_c$ contributes little to the learned feature representation when $\gamma_c$ is small. Hence, the convolution in Eqn.(\ref{eq:conv-norm-relu}) can be discarded safely, and filter $\mathbf{w}_c$ can thus be pruned. 
%Unlike these methods that measure the importance of channels by their magnitude of norm, we attempt to develop a probability-based method to represent the channel importance in the model.
Unlike these criterion-based methods that
deterministically prune unimportant filters and rely on a heuristic pruning procedure as shown in Fig.\ref{fig:Fig1}(a), we explore a probability-based channel pruning framework where less important channels are pruned in a stochastic manner.
%\vspace{-2pt}

\textbf{Activation probability.} To this end, we define an activation probability of a channel by exploring the property of the BN layer. Those channels with a larger activation probability could be preserved with a higher probability.  To be specific, since $\bar{\mathbf{x}}_c$ is acquired by subtracting the sample mean and being divided by the sample variance, we can treat each channel feature as a random variable following standard Normal distribution \cite{arpit2016normalization}, denoted as ${\bar{{X}}}_c$. Note that only positive parts can be activated by ReLU function. Proposition \ref{remark:1} gives the activation probability of the $c$-th channel, \ie $P({\tilde{{X}}}_c)>0$.
\vspace{-3pt}
\begin{prop}\label{remark:1}
Let a random variable ${\bar{{X}}}_c \sim\mathcal{N}(0,1)$ and $Y_c=max\{0,\gamma_c{\bar{{X}}}_c+\beta_c\}$. Then we have  (1) $P(Y_c>0) =P({\tilde{{X}}}_c>0)= (1+\mathrm{Erf}(\beta_c/(\sqrt{2}|\gamma_c|))/2$ where $\mathrm{Erf}(x)=\int_0^x2/\sqrt{\pi} \cdot \mathrm{exp}^{-t^2}dt$, and (2) $P(\tilde{{X}}_c>0) = 0 \Leftrightarrow {\beta}_c \leq 0\, \mathrm{and}\, {{\gamma}_c\rightarrow 0}$.
%(2)\,$\mathbb{E}_z[y]=\frac{1}{\sqrt{2\pi}}$ and $\mathbb{E}_z[y^2]=\frac{\pi-1}{2\pi}$ if $\gamma_c=1$ and $\beta_c=0$.
\end{prop}
\vspace{-0.1in}
%Therefore, we have ${\tilde{{X}}}_c \sim \mathcal{N}(\beta_c, \gamma_c^2)$ by Eqn.(\ref{eq:conv-norm-relu}). Note that only positive parts can be activated by ReLU activation; the activation probability of the $c$-th channel can be defined by
%
%\vspace{-0.1in}
%\begin{equation}\label{eq:activation-prob}
%P(\tilde{{X}}_c>0) = \int^{+\infty}_{0}\frac{1}{\sqrt{2\pi\gamma_c^2}}\mathrm{exp}^{-\frac{(x-\beta_c)^2}{2\gamma_c^2}}
%\end{equation}
%\vspace{-0.1in}
%
%With Eqn.(\ref{eq:activation-prob}),  
Note that a pruned channel can be modelled by $P(\tilde{{X}}_c>0) = 0$. With Proposition \ref{remark:1} (see proof in Appendix \ref{sec:A.2}), we know that the unnecessary channels satisfy that $\gamma_c$ approaches $0$ and $\beta_c$ is negative. To achieve channel pruning, previous compression techniques \cite{li2017, zhao2019variational}  merely impose a regularization on $\gamma_c$, which would deteriorate the representation power of unpruned channels \cite{perez2018film, wang2020deep}. Instead, we adopt the idea of whitening to build a probabilistic channel pruning framework where unnecessary channels are stochastically regarded with a small activation probability while important channels are preserved with a large activation probability.
%Instead of deterministically zeroing out unimportant channels that harms learning capacity of the network as norm-based methods \cite{liu2017learning,he2018soft} did, we develop 
%We have assigned each channel with an activation probability, the following section presents a probabilistic channel pruning framework where less important channels are identified by utilizing activation probability in Proposition \ref{remark:1}.

%We have assigned each channel with an activation probability, the following section presents how to easily identify unimportant channels by adjusting their activation probabilities.

\section{Batch Whitening Channel Pruning}
\begin{figure}
\begin{center}
\includegraphics[scale=0.6]{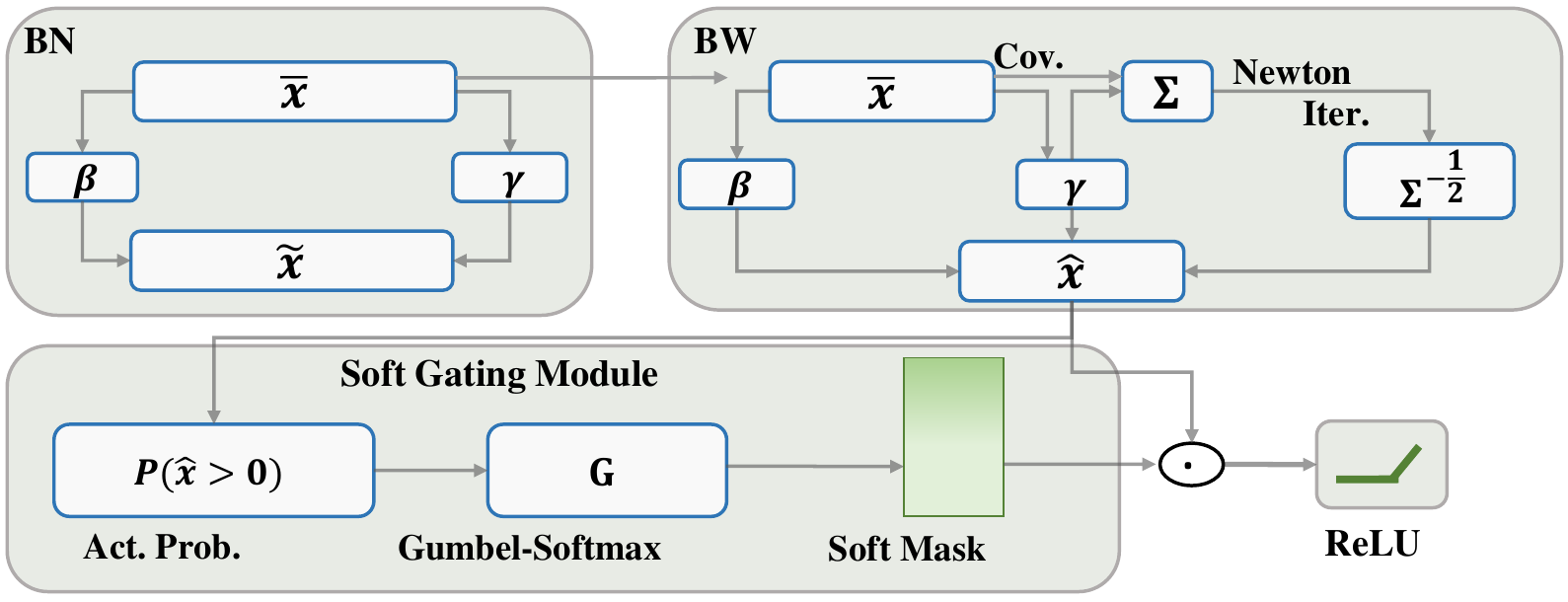}
\end{center}
\vspace{-0.2in}
  \caption{A schematic of the proposed Batch Whitening Channel Pruning (BWCP) algorithm  that consists of a BW module and a soft sampling procedure. By modifying BN layer with a whitening operator, the proposed BW technique adjusts activation probabilities of different channels. These activation probabilities are then utilized by a soft sampling procedure.}
\label{fig:BW-pipeline}
\vspace{-0.2in}
\end{figure}
\vspace{-0.05in}
This section introduces the proposed batch whitening channel pruning (BWCP) algorithm, which contains a batch whitening  module  that can adjust the activation probability of channels, and a soft sampling module that stochastically prunes channels with the activation probability adjusted by BW. The whole pipeline of BWCP is illustrated in Fig.\ref{fig:BW-pipeline}. 
%In principle, BWCP operates after the convolution layer and before ReLU activation.  

By modifying the BN layer in  Eqn.(\ref{eq:conv-norm-relu}), we have the formulation of BWCP,
\vspace{-0.1in}
\begin{equation}\label{eq:BWCP}
{\mathbf{x}}_{c}^{\mathrm{out}} = \underbrace{\hat{\mathbf{x}}_{c}}_{\mathrm{batch\,  whitening}} \odot\quad \underbrace{{m}_c(P(\hat{{X}}_c>0))}_{\mathrm{soft\, sampling}}
\end{equation}
%\vspace{-0.1in}
where $\mathbf{x}_{c}^{out}, \hat{\mathbf{x}}_{c} \in\mathbb{R}^{ N\times H  \times W}$ denote the output of proposed BWCP algorithm and BW module, respectively. `$\odot$' denotes broadcast multiplication.  $m_c\in [0,1]$ denotes a soft sampling that takes the activation probability of output features of BW (\ie $P(\hat{{X}}_c>0)$) and returns a soft mask.  The closer the activation probability is to $0$ or $1$, the more likely the mask is to be hard. To distinguish important channels from unimportant ones, BW is proposed to increase the activation probability of useful channels while keeping the probability of unimportant channels unchanged during training.
%When the mask becomes hard binary variables, BWCP can turn on/off the corresponding channel. 
%
Since Eqn.(\ref{eq:BWCP}) always  retain all channels in the network, our BWCP can preserve the learning capacity of the original network during training \cite{he2018soft}. The following sections present BW and soft sampling module in detail.

\subsection{Batch Whitening}\label{sec:BW}
Unlike previous works \cite{zhao2019variational, kang2020operation} that simply measure the importance of channels by parameters in BN layer, we attempt to whiten features after BN layer by the proposed BW module. We show that BW can change the activation probability of channels according to their importances without adding additional parameters and computational overhead in inference. 

As shown in Fig.\ref{fig:BW-pipeline}, 
%This section introduces a generally applicable tool batch whitening (BW) %It can increase the probability of useful channels but reduce the activation probability of unimportant channels. 
%which 
BW acts after the BN layer. %The output of BW in Eqn.(\ref{eq:BWCP}) $\hat{\mathbf{x}}_{nij}$ is formulated by
By rewriting Eqn.(\ref{eq:conv-norm-relu}) into a  vector form, we have the formulation of BW, 
\begin{equation}\label{eq:BW-gb}
\hat{\mathbf{x}}_{nij} = \Sigmainv(\boldsymbol{\gamma} \odot \bar{\mathbf{x}}_{nij} + \boldsymbol{\beta})
\end{equation}
where $\hat{\mathbf{x}}_{nij}\in\mathbb{R}^{ C\times1}$ is a vector of $C$ elements that denote the output of BW for the $n$-th sample at location $(i,j)$ for all channels. $\Sigmainv$ is a whitening operator and $\mathbf{\Sigma}\in\R^{C\times C}$ is the covariance matrix of channel features $\{\tilde{\mathbf{x}}_c\}_{c=1}^C$. 
Moreover, $\boldsymbol{\gamma}\in \R^{C\times1}$ and $\boldsymbol{\beta} \in \R^{C\times1}$ are two vectors by stacking $\gamma_c$ and $\beta_c$ of all the channels respectively. $\bar{\mathbf{x}}_{nij}\in \R^{C\times1}$ is a vector by stacking elements from all channels of $\bar{x}_{ncij}$ into a column vector. 
\textbf{Training and inference.} Note that BW in Eqn.(\ref{eq:BW-gb}) requires computing a root inverse of a covariance matrix of channel features after the BN layer. Towards this end, we calculate the covariance matrix $\mathbf{\Sigma}$ within a batch of samples during each training step as given by
\vspace{-0.1in}
\begin{equation}\label{eq:covar-BW}
\mathbf{\Sigma}= \frac{1}{NHW}\sum_{n,i,j=1}^{N,H,W} (\boldsymbol{\gamma} \odot \bar{\mathbf{x}}_{nij})(\boldsymbol{\gamma} \odot \bar{\mathbf{x}}_{nij})\tran=(\boldsymbol{\gamma}\boldsymbol{\gamma}\tran)\odot\boldsymbol{\rho}
\end{equation}
%
%\vspace{-0.1in}
where $\boldsymbol{\rho}$ is a C-by-C correlation matrix of channel features $\{\bar{\mathbf{x}}_c\}_{c=1}^C$ (see details in Appendix \ref{sec:A.1}).
%$\mathbf{\rho}=\frac{1}{NHW}\sum_{n,i,j=1}^{N,H,W}\bar{\mathbf{x}}_{nij}\bar{\mathbf{x}}_{nij}\tran$ is a C-by-C matrix that denotes the correlation matrix of channels features $\{\bar{\mathbf{x}}_c\}_{c=1}^C$. 
The Newton Iteration is further employed to calculate its root inverse, $\Sigmainv$, as given by the following iterations
\vspace{-0.1in}
\begin{equation}\label{eq:Newtoniter}
\mathbf{\Sigma}_k=\frac{1}{2}(3\mathbf{\Sigma}_{k-1}-\mathbf{\Sigma}_{k-1}^3\mathbf{\Sigma}),\, k=1,2,\cdots,T.
\end{equation}
\vspace{-2pt}
%
\iffalse
\vspace{-2pt}
\begin{equation}\label{eq:Newtoniter}
\left\{
\begin{array}{l}
\mathbf{\Sigma}_0=\mathbf{I}\\
\mathbf{\Sigma}_k=\frac{1}{2}(3\mathbf{\Sigma}_{k-1}-\mathbf{\Sigma}_{k-1}^3\mathbf{\Sigma}),\, k=1,2,\cdots,T.
\end{array}
\right.
\end{equation}
\vspace{-1pt}
\fi
where $k$ and $T$ are the iteration index and iteration number respectively and $\mathbf{\Sigma}_0 = \mathbf{I}$ is a identity matrix. 
%Note that the convergence of Eqn.(\ref{eq:Newtoniter}) is guaranteed if $\left\|\mathbf{I}-\mathbf{\Sigma}\right\|_2<1$ \cite{bini2005algorithms}.
Note that when $\left\|\mathbf{I}-\mathbf{\Sigma}\right\|_2<1$, Eqn.(\ref{eq:Newtoniter}) converges to $\mathbf{\Sigma}^{-\frac{1}{2}}$ \cite{bini2005algorithms}.
To satisfy this condition, $\mathbf{\Sigma}$ can be normalized by $\mathbf{\Sigma}/\mathrm{tr}(\mathbf{\Sigma})$ following  \cite{huang2019iterative}, where $\mathrm{tr}(\cdot)$ is the trace operator. In this way, the normalized covariance matrix can be written as $\mathbf{\Sigma}_N=\boldsymbol{\gamma}\boldsymbol{\gamma}\tran\odot \boldsymbol{\rho}/\left\|\boldsymbol{\gamma}\right\|_2^2$.
%$\mathbf{\Sigma}=\frac{\mathbf{\gamma}\mathbf{\gamma}\tran}{\left\|\mathbf{\gamma}\right\|_2^2} \odot \mathbf{\rho}$.

During inference,  we use the moving average to calculate the population estimate of $\hat{\mathbf{\Sigma}}_N^{-\frac{1}{2}}$  by following the updating rules, $\hat{\mathbf{\Sigma}}_N^{-\frac{1}{2}}=(1-g)\hat{\mathbf{\Sigma}}^{-\frac{1}{2}}_N+g\mathbf{\Sigma}^{-\frac{1}{2}}_N$.
%
%\begin{equation}\label{eq:movinginfernce}
    %\hat{\mathbf{\Sigma}}^{-\frac{1}{2}}=(1-g)\hat{\mathbf{\Sigma}}^{-\frac{1}{2}}+g\mathbf{\Sigma}^{-\frac{1}{2}}
%\end{equation}
Here $\mathbf{\Sigma}_N$ is the covariance calculated within each mini-batch at each training step, and $g$ denotes the momentum of moving average. Note that $\hat{\mathbf{\Sigma}}^{-\frac{1}{2}}_N$ is fixed during inference, the proposed BW does not introduce extra costs in memory or computation since $\hat{\mathbf{\Sigma}}^{-\frac{1}{2}}_N$ can be viewed as a convolution kernel with size of $1$, which can be absorbed into previous convolutional layer.

%we can just substitute $\mathbf{\gamma}$ and $\mathbf{\beta}$ in BN layers with $\hat{\mathbf{\Sigma}}^{-\frac{1}{2}}{\mathbf{\gamma}}$ and $\hat{\mathbf{\Sigma}}^{-\frac{1}{2}}{\mathbf{\beta}}$ respectively.

\iffalse
\begin{figure}
\begin{center}
\includegraphics[scale=0.43]{./probcases-1.pdf}
\end{center}
\vspace{-0.1in}
  \caption{Illustration of how our proposed BW changes the activation probability in three cases, \ie $\beta_c \ll 0$, $\beta_c\leq 0 $ and $\beta_c \gg 0$ respectively. \textbf{(a)} shows BW reduce activation probability of unimportant channels by lowering the value of $\beta$ when $\beta_c \ll 0$. \textbf{(b)} illustrates that BW enlarges the magnitude of $\gamma$ to increase activation probability of slightly important channels when $\beta_c\leq 0 $. and BW in \textbf{(c)} $\beta_c \gg 0$ increases the value of $\beta$ to enhance activation probability of important channels. }
\label{fig:BW-probcases}
\end{figure}
\fi
%

\subsection{Analysis of BWCP}\label{sec:BW-analysis}
In this section, we show that BWCP can easily identify unimportant channels by increasing the difference of activation between important and unimportant channels.
%can enhance the activation probability of useful channels, while reducing the probability of less important channels through the proposed BW technique.
%
\vspace{-0.1in}
\begin{prop}\label{remark:2}
%Given Eqn.(\ref{eq:BW-gb}) and Eqn.(\ref{eq:conv-norm-relu}), suppose that for all $c\in[C]$, $(\bar{\mathbf{x}}_{nij})_c$ corresponds to a identical R.V. following standard Normal, 
Let a random variable ${\bar{{X}}} \sim\mathcal{N}(0,1)$ and $Y_c=max\{0,[\hat{\mathbf{\Sigma}}^{-\frac{1}{2}}_N (\boldsymbol{\gamma} \odot \bar{{X}}+\boldsymbol{\beta})]_c\}$. Then we have  $P(Y_c>\delta) =P({\hat{{X}}}_c>\delta)= (1+\mathrm{Erf}((\hat{\beta}_c-\delta)/(\sqrt{2}|\hat{\gamma}_c|))/2$, where $\delta$ is a small positive constant, $\hat{\gamma}_c$  and $\hat{\beta}_c$ are two equivalent scale parameter and bias defined by BW module. Take $T=1$ in Eqn.(\ref{eq:Newtoniter}) as an example, we have $\hat{\gamma}_c =\frac{1}{2}(3\gamma_c-\sum_{d=1}^{C}{\gamma_d^2\gamma_c\rho_{dc}/\left\|\boldsymbol{\gamma}\right\|_2^2})$, and $\hat{\beta}_c =\frac{1}{2}(3\beta_c-\sum_{d=1}^{C}{\beta_d\gamma_d\gamma_c\rho_{dc} /\left\|\boldsymbol{\gamma}\right\|_2^2})$
%
\iffalse
\begin{align}
\hat{\gamma}_c &=\frac{1}{2}(3\gamma_c-(\sum_{d=1}^{C}\frac{\gamma_d^2\rho_{dc}}{\left\|\boldsymbol{\gamma}\right\|_2^2})\gamma_c),\label{eq:equgamma}
\hat{\beta}_c &=\frac{1}{2}(3\beta_c-(\sum_{d=1}^{C}\frac{\beta_d\gamma_d\rho_{dc}}{\left\|\boldsymbol{\gamma}\right\|_2^2})\gamma_c), \label{eq:equbeta}
\end{align}
\fi
%
where $\rho_{dc}$ is the Pearson's correlation between channel features $\bar{\mathbf{x}}_c$ and $\bar{\mathbf{x}}_d$. 
\vspace{-0.05in}
\end{prop}
%By comparing Eqn.(\ref{eq:BW-gb}) with Eqn.(\ref{eq:conv-norm-relu}), we see that 
%
%
\iffalse
\begin{figure}
\begin{center}
\includegraphics[scale=0.27]{./probcases-2.pdf}
\end{center}
\vspace{-2pt}
  \caption{Illustration of how our proposed BW changes the activation probability in two cases, \ie $\beta_c \ll 0$ and $\beta_c \gg 0$ respectively. \textbf{(a)} shows BW reduce activation probability of unimportant channels by lowering the value of $\beta$ when $\beta_c \ll 0$ while \textbf{(b)} shows that BW increases the value of $\beta$ to enhance activation probability of important channels when $\beta_c \gg 0$. }
\label{fig:BW-probcases}
\vspace{-0.2in}
\end{figure}
\fi
By Proposition .\ref{remark:2}, BWCP can adjust activation probability by changing the values of $\gamma_c$ and $\beta_c$ in Proposition \ref{remark:1} through BW module (see detail in Appendix \ref{sec:A.3}). Here we introduce a small positive constant $\delta$ to avoid the small activation feature value.
To see how BW changes the activation probability of different channels, we consider two cases as shown in Proposition \ref{remark:3}.

\textbf{Case 1:} $\beta_c\leq 0 $ and $\gamma_c\rightarrow 0$. In this case, the $c$-th channel of the BN layer would be activated with a small activation probability as it sufficiently approaches zero. We can see from Proposition \ref{remark:3}, the activation probability of $c$-th channel is reduced after BW is applied in this case, showing that the proposed BW module can keep the unimportant channels unchanged. 
\textbf{Case 2:} $|\gamma_c|>0$. For this case, the $c$-th channel of the BN layer would be activated with a high activation probability.   We can see from Proposition \ref{remark:3} whose detailed proof is provided in Appendix \ref{sec:A.4}, the activation probability of $c$-th channel is enlarged after BW is applied. Therefore, our proposed BW module can increase the activation probability of important channels. Detailed proof of Proposition \ref{remark:3} can be found in Appendix. We also empirically verify Proposition \ref{remark:3} in Sec. \ref{sec:ablation}.
Notice that we neglect a trivial case in which the channel can be also activated (\ie $\beta_c >0 $ and $|\gamma_c|\rightarrow 0$). In fact, the channels in this case can be removed because it can be deemed as a bias term of the convolution.

%
%
%
%To sum up, our proposed BWCP can not only better identify unimportant channels by reducing their activation probabilities but also enhance the representation ability of useful channels by increasing their activation probabilities.
%
\subsection{Soft Sampling Module}
%Although the proposed BW with a sparse regularization can give unimportant channels a smaller activation probability,  abruptly pruning them after training would result in a performance drop. To alleviate this, we 
%BWCP maps the output of the proposed BW through a 
The soft sampling procedure samples the output of BW through a set of differentiable masks. To be specific, as shown in Fig.\ref{fig:BW-pipeline}, we leverage the Gumbel-Softmax sampling \cite{jang2017categorical} that takes the activation probability generated by BW and produces a soft mask as given by 
\vspace{-0.05in}
\begin{equation}\label{eq:soft-gate}
    m_c = \mathrm{GumbelSoftmax}(P(\hat{{X}}_c>0);\,\tau)
\end{equation}
%\vspace{-0.05in}
where $\tau$ is the temperature. By Eqn.(\ref{eq:BWCP}) and Eqn.(\ref{eq:soft-gate}), BWCP stochastically prunes unimportant channels with activation probability. A smaller activation probability makes $m_c$ more likely to be close to $0$. Hence, our proposed BW can help identify less important channels by enlarging the activation probability of important channels, as mentioned in Sec.\ref{sec:BW-analysis}. Note that $m_c$ can converge to $0$-$1$ mask when $\tau$ approaches to $0$. In the experiment, we find that setting $\tau=0.5$ is enough for BWCP to achieve hard pruning at test time. 

\begin{prop}\label{remark:3}
Let $\delta=\left\| \boldsymbol{\gamma}\right\|_2\sqrt{\sum_{j=1}^C({\gamma}_j\beta_c-\gamma_c{\beta}_j)^2\rho_{cj}^2}/(\left\| \boldsymbol{\gamma}\right\|_2^2-\sum_{j=1}^C\gamma_j^2\rho_{cj})$.  With $\hat{\gamma}$ and $\hat{\beta}$ defined in Proposition \ref{remark:2}, we have (1) $P({\hat{{X}}}_c > \delta) = 0$ if $|\gamma_c|\rightarrow 0$ and $\beta_c\leq 0$, and (2) $P({\hat{{X}}}_c > \delta) \geq P({\tilde{{X}}}_c \geq \delta)$ if $|\gamma_c| > 0$.
\end{prop}

Note that  the number of channels in the last convolution layer must be the same as previous blocks due to the element-wise summation in the recent advanced CNN architectures \cite{he2016deep,huang2017densely}. We solve this problem by letting BW layer in the last convolution layer and shortcut share the same mask as discussed in Appendix \ref{sec:A.5}.
%Since Eqn.(\ref{eq:BWCP}-\ref{eq:BW-gb}) and Eqn.(\ref{eq:soft-gate}) that represent forward propagation of BWCP defines differentiable transformations, our proposed can train and prune the model in an end-to-end manner.

\iffalse
\textbf{Solution to residual issue.} The recent advanced CNN architectures usually have residual blocks with shortcut connections \cite{he2016deep,huang2017densely}. As shown in Fig.2 in Appendix \ref{sec:A.4}, when pruning channels in a residual block, the number of channels in the last convolution layer must be the same as previous blocks due to the element-wise summation. We solve this problem by letting BW layer in the last convolution layer and shortcut share the same mask. Specifically, we acquire the sharing mask by a simple product of their respective masks, as given by $m_c = m_c^{last} \odot {m}_c^s $ where ${m}_c^{last}$ and ${m}_c^s$ denote masks in soft sampling module of the last convolution layer and shortcut. In doing so, their activated output channels must be the same.
\fi

\subsection{Training of BWCP}\label{sec:training-dmcp}
This section introduces a sparsity regularization, which makes the model compact, and then describes the training algorithm of BWCP.

%
%
\iffalse
\vspace{-2pt}
\begin{equation}\label{eq:criterion-prob}
P(\tilde{{X}}_c>0) = 0 \Leftrightarrow \beta_c \leq 0\, \mathrm{and}\, {\gamma_c=0}
\end{equation}
\vspace{-2pt}
\fi
%
\textbf{Sparse Regularization.} We can see from Proposition.\ref{remark:1} that the main characteristic of pruned channels in BN layer is that $\gamma_c$ sufficiently approaches $0$, and $\beta_c$ is negative. By Proposition \ref{remark:3}, we find that it is also a necessary condition that a channel can be pruned after BW module is applied. Hence, we obtain unnecessary channels by directly imposing a regularization on $\gamma_c$ and $\beta_c$ as given by 
%
%
\iffalse
Note that a pruned channel can be modelled by $P(\tilde{{X}}_c>0) = 0$. By Proposition \ref{prop:spare-regu}, we can acquire a necessary  condition that a channel can be pruned.
%we can acquire a necessary  condition that a channel can be pruned 
%with probability $1$ as given by $P(\tilde{{X}}_c>0) = 0 \Leftrightarrow \hat{\beta}_c \leq 0\, \mathrm{and}\, {\hat{\gamma}_c=0}$.
%
Detailed proofs are provided in Appendix \ref{sec:A.5}. The condition shows the main characteristic of pruned channels, \ie $\gamma_c$ sufficiently approaches $0$, and $\beta_c$ is negative, which motivates us to obtain unnecessary channels by imposing a regularization on $\gamma_c$ and $\beta_c$ as given by 
\fi
%
\begin{equation}\label{eq:sparse_loss}
\mathcal{L}_{\mathrm{sparse}} = \sum\nolimits_{c=1}^C\lambda_1 |\gamma_c| + \lambda_2\beta_c
\end{equation}
where the first term makes $\gamma_c$ small, and the second term encourages $\beta_c$ to be negative.  The above sparse regularizer is imposed on all BN layers of the network. By changing the strength of regularization (\ie $\lambda_1$ and $\lambda_2$), we can achieve different pruning ratios.

\iffalse
\vspace{-0.05in}
\begin{prop}\label{prop:spare-regu}
Let a random variable ${\bar{{X}}}_c \sim\mathcal{N}(0,1)$ and $Y_c=max\{0, \hat{\gamma}_c{\bar{{X}}}_c+\hat{\beta}_c\}$ where $\hat{\gamma}_c$ and $\hat{\beta}_c$ are defined in Proposition \ref{remark:2}. Then we have ${\beta}_c \leq 0\, \mathrm{and}\, {{\gamma}_c=0} \Rightarrow P(\hat{{X}}_c>0) = 0$.
\end{prop}
\vspace{-0.05in}
\fi
%\textbf{Soft Gating module.} 
%

\textbf{Training Algorithm.} BWCP can be easily plugged into a CNN by modifying the traditional BN operations. Hence, the training of BWCP can be simply implemented in existing software platforms such as PyTorch and TensorFlow. In other words, the forward propagation of BWCP can be represented by Eqn.(\ref{eq:BWCP}-\ref{eq:BW-gb}) and Eqn.(\ref{eq:soft-gate}), all of which  define differentiable transformations. Therefore, our proposed BWCP can train and prune deep models in an end-to-end manner. Appendix \ref{sec:A.6} also provides the explicit gradient back-propagation of BWCP. On the other hand, we do not introduce extra parameters to learn the pruning mask ${m}_c$. Instead, ${m}_c$ in  Eqn.(\ref{eq:soft-gate}) is totally determined by the parameters in BN layers including $\boldsymbol{\gamma}, \boldsymbol{\beta}$ and $\mathbf{\Sigma}$. Hence, we can perform joint training of pruning mask  ${m}_c$ and model parameters. %Furthermore, , facilitating the implementation of BWCP in the platforms without auto differentiation. 
The BWCP framework is provided in Algorithm 1 of Appendix Sec A.6

After training, we obtain the final compact model by directly pruning channels with a mask value of $0$. Since the pruning mask ${m}_c$ becomes hard $0$-$1$ binary variables, our proposed BWCP does not need an extra fine-tuning procedure.

%Pruning mask $\mathbf{m}$ in Eqn.(\ref{eq:sparse_loss}) together with Eqn.(\ref{eq:BW-mask}) enable the proposed BWCP to train and prune from scratch.  On the one hand, it is obvious that the sparse regularization in Eqn.(\ref{eq:sparse_loss}) is differentiable \textit{w.r.t.} parameters in BN layers. On the other hand,  we do not introduce extra parameters to learn the pruning mask $\mathbf{m}$. Instead, $\mathbf{m}$ in  Eqn.(\ref{eq:BW-mask}) is totally determined by the parameters in BN layers including $\mathbf{\gamma}, \mathbf{\beta}$ and $\mathbf{\Sigma}$.
%where $\mathbf{\Sigma}$ is a function of $\mathbf{\gamma}, \mathbf{w}$ and layer input $\mathbf{z}$. 
%Since Gumbel-Softmax in the soft gating module is differentiable \textit{w.r.t.} input probability logits, the backward propagation for the pruning mask can be calculated through Eqn.(\ref{eq:conv-norm-relu}-\ref{eq:Newtoniter}) and Eqn.(\ref{eq:BW-mask}), \ie $\frac{\partial \mathbf{m}}{\partial \mathbf{\gamma}}, \frac{\partial \mathbf{m}}{\partial \mathbf{\beta}}$ and $\frac{\partial \mathbf{m}}{\partial \mathbf{w}}$ whose detailed derivations are provided in Appendix.
%

%-------------------------------------------------------------------------

%------------------------------------------------------------------------
\section{Experiments}

In this section, we extensively experiment with the proposed BWCP. We show the advantages of BWCP in both recognition performance and FLOPs reduction comparing with existing channel pruning methods. We also provide an ablation study to analyze the proposed framework.
\begin{table*}[!ht]
	\centering
	\caption{
		Performance comparison between our proposed approach BWCP and other methods on CIFAR-10. 
		%``\cmark'' and ``\xmark'' in the Fine-tune column indicate whether need an extra fine-tuning procedure or not. 
		``Baseline Acc." and ``Acc." denote the accuracies of the original and pruned models, respectively.	``Acc. Drop" means the accuracy of the base model minus that of pruned models (smaller is better).	``Channels $\downarrow$", ``Model Size $\downarrow$", and ``FLOPs $\downarrow$" denote the relative reductions in individual metrics compared to the unpruned networks (larger is better). `*' indicates the method needs a extra fine-tuning to recover performance. The best-performing results are highlighted in bold.
	}
	%\vspace{-0.1in}
	\scalebox{0.65}{
		\begin{tabular}{c c c c c c c c c}		
			\toprule
			\multirow{1}{*}{Model} & Method  & Baseline Acc. (\%) & Acc. (\%) & Acc. Drop & Channels $\downarrow$ (\%) & Model Size $\downarrow$ (\%) & FLOPs $\downarrow$ (\%) \\
			\hline %\hline
			\multirow{5}{*}{ResNet-56}  
			&DCP*~\cite{zhuang2018discrimination} & 93.80 & 93.49 & 0.31 & -  & 49.24  & 50.25 \\
			&AMC*~\cite{he2018amc} & 92.80 & 91.90 & 0.90 & -  & -  & 50.00 \\
			&SFP~\cite{he2018soft}    & 93.59 & 92.26 & 1.33 & 40  & --  & \textbf{52.60} \\
			 & FPGM~\cite{he2019filter} & 93.59 & 92.93 & 0.66 & 40  & --  & \textbf{52.60} \\
			 & SCP~\cite{kang2020operation}  &  93.69 & 93.23 & 0.46 & \textbf{45} & \textbf{46.47} &  51.20 \\
			 & BWCP (Ours)   & 93.64 & 93.37 & \textbf{0.27} & 40 & 44.42 &  50.35 \\
			\hline %\hline
				\multirow{4}{*}{DenseNet-40} 
			&Slimming*~\cite{liu2017learning} & 94.39 & 92.59 & 1.80 & 80  & 73.53  & 68.95 \\			
			& Variational Pruning~\cite{zhao2019variational}  & 94.11 & 93.16 &0.95&  60 & 59.76 & 44.78 \\
			%\cdashline{2-9}
			%Slimming~\cite{liu2017learning} & DenseNet-40&X& 94.39 & 12.78 & 81.61 & 80  & 73.53  & 68.95 \\
			 &SCP~\cite{kang2020operation}  & 94.39 & 93.77 & 0.62 & 81 & 75.41 & 70.77 \\
			%SCP~\cite{kang2020operation} & DenseNet-40  & X & 94.39 & 93.77 & 0.62 & \textbf{81} & \textbf{75.41} & \textbf{70.77} \\ 
			 & BWCP (Ours)   & 94.21 & 93.82 & \textbf{0.39} & \textbf{82} & \textbf{76.03} & \textbf{71.72} \\
			\hline %\hline
			
			\multirow{4}{*}{VGGNet-16}
			&Slimming*~\cite{liu2017learning} & 93.85 & 92.91 & 0.94 & 70  & 87.97  & 48.12 \\	
			&Variational Pruning~\cite{zhao2019variational}  & 93.25 & 93.18 & 0.07 & 62 &73.34 & 39.10 \\
			%\cdashline{2-9}
			%Slimming~\cite{liu2017learning}& VGGNet-16&X& 93.85 & 10.00 & 83.85 & 70 & 87.97 & 48.12 \\		
			%&Variational Pruning~\cite{zhao2019variational}  & 93.25 & 93.18 & 0.07 & 62 &73.34 & 39.10 \\
			&SCP~\cite{kang2020operation}  & 93.85 & 93.79 & 0.06 & {75} & {93.05} & {66.23} \\
			 & BWCP (Ours)  &  93.85 & 93.82 & \textbf{0.03} & \textbf{76}& \textbf{93.12}&\textbf{68.08}   \\
			\bottomrule
		\end{tabular}
	}
	\label{table:cifar10_table}
	\vspace{-0.2in}
\end{table*}

\subsection{Datasets and Implementation Settings}
We evaluate the performance of our proposed BWCP on various image classification benchmarks, including CIFAR10/100 \cite{krizhevsky2009learning} and ImageNet \cite{russakovsky2015imagenet}. The CIFAR-10 and CIFAR-100 datasets have $10$ and $100$ categories, respectively, while both contain $60$k color images with a size of $32\times 32$, in which $50$k training images and $10$K test images are included. Moreover, the ImageNet dataset consists of $1.28$M training images and $50$k validation images. Top-1 accuracy are used to evaluate the recognition performance of models on CIFAR 10/100. Top-1 and Top-5 accuracies are reported on ImageNet. We utilize the common protocols, \ie number of parameters and Float Points Operations (FLOPs) to obtain model size and computational consumption.

For CIFAR-10/100,  we use ResNet~\cite{he2016deep}, DenseNet~\cite{huang2017densely}, and VGGNet~\cite{simonyan2014very} as our base model. For ImageNet, we use ResNet-34 and ResNet-50. We compare our algorithm with  other channel pruning methods without a fine-tuning procedure. Note that a extra fine-tuning process would lead to remarkable improvement of performace \cite{ye2020good}. For fair comparison, we also fine-tune our BWCP to compare with those pruning methods that requires a fine-tuning as did in FPGM \cite{he2019filter} and SCP \cite{kang2020operation}. The training configurations are provided in Appendix \ref{sec:B.1}. The base networks and BWCP are trained together from scratch for all of our models.

 %ResNet~\cite{he2016deep}, DenseNet~\cite{huang2017densely}, and VGGNet~\cite{simonyan2015very} which are widely used for image classification tasks. 
%We compare our approach with Slimming~\cite{liu2017learning}, Variational Pruning~\cite{zhao2019variational} and SCP~\cite{kang2020operation}, which prunes redundant channels by making use of the BN layers. We also compare the proposed method with SFP~\cite{he2018soft} and FPGM~\cite{he2019filter} for CIFAR-10, which do not need the fine-tuning process same as our approach. 

%The results of Slimming are given by our reproduction from TensorFlow implementation.

%-------------------------------------------------------------------------
\subsection{Results on CIFAR-10 }\label{sec:results_cifar}
For CIFAR-10 dataset, we evaluate our BWCP on ResNet-56, DenseNet-40 and VGG-16 and compare our approach with Slimming~\cite{liu2017learning}, Variational Pruning~\cite{zhao2019variational} and SCP~\cite{kang2020operation}. These methods prune redundant channels using BN layers like our algorithm. We also compare BWCP with previous strong baselines sych as AMC \cite{he2018amc} and DCP \cite{zhuang2018discrimination}. The results of slimming are obtained from SCP \cite{kang2020operation}. 
As mentioned in Sec.\ref{sec:BW-analysis}, our BWCP adjusts their activation probability of different channels. Therefore, it is expected to present better recognition accuracy with comparable computation consumption by entirely exploiting important channels.
As shown in Table~\ref{table:cifar10_table}, our BWCP achieves the lowest accuracy drops and comparable FLOPs reduction compared with existing channel pruning methods in all tested base networks. 
For example, although our model is not fine-tuned, the accuracy drop of the pruned network given by BWCP based on DenseNet-40 and VGG-16 outperforms Slimming with fine-tuning by $1.41$\% and $0.91$\% points, respectively. And ResNet-56 pruned by BWCP attains better classification accuracy without an extra fine-tuning stage.
Besides, our method achieves superior accuracy compared to the Variational Pruning even with significantly smaller model sizes on DensNet-40 and VGGNet-16. Note that BWCP can slightly outperform the recently proposed SCP based on ResNet-56  by $0.19$\% points, demonstrating its effectiveness. We also report results of BWCP on CIFAR100 in Appendix \ref{sec:B.2}.

\subsection{Results on ImageNet}

For ImageNet dataset, we test our proposed BWCP on two representative base models ResNet-34 and ResNet-50. The proposed BWCP is compared with SFP~\cite{he2018soft}), FPGM~\cite{he2019filter}, SSS \cite{huang2018data}),  SCP~\cite{kang2020operation} HRank \cite{lin2020hrank} and DSA~\cite{ning2020dsa} since they prune channels without an extra fine-tuning stage. As shown in Table \ref{table:ImageNet_table}, we see that BWCP consistently outperforms its counterparts in recognition accuracy under comparable FLOPs. 
For ResNet-34, FPGM ~\cite{he2019filter} and SFP~\cite{he2018soft} without
fine-tuning accelerates ResNet-34 by $41.1$\% speedup ratio
with $2.13$\% and $2.09$\% accuracy drop respectively, but our BWCP without finetuning achieve almost the same speedup ratio with only $1.27$\% top-1 accuracy drop. On the other hand, BWCP also significantly outperforms FPGM ~\cite{he2019filter} by  $0.84$\% top-1 accuracy after going through a fine-tuning stage.
For ResNet-50, BWCP still achieves better performance compared with other approaches. For instance, at the level of $40$\% FLOPs reduction, the top-1 accuracy of BWCP  exceeds SSS~\cite{huang2018data} by $3.60$\%.
Moreover, BWCP outperforms  DSA~\cite{ning2020dsa} by top-1 accuracy of $0.22$\% and  $0.13$\% at level of $40$\% and $50$\%  FLOPs  respectively even DSA requires extra $20$ epoch warm-up. However, 
 BWCP has slightly lower top-5 accuracy than DSA~\cite{ning2020dsa}.
\begin{table*}[!t]
	\centering
	\caption{
		Performance of our proposed BWCP and other pruning methods on ImageNet using  base models ResNet-34 and ResNet-50. '*' indicates the pruned model is fine-tuned.
	}
	%\vspace{-0.1in}
	\scalebox{0.70}{
		\begin{tabular}{c c  c c c c c}
			\toprule
			Model & Method  & Baseline Top-1 Acc. (\%) & Baseline Top-5 Acc. (\%) & Top-1 Acc. Drop & Top-5 Acc. Drop & FLOPs $\downarrow$ (\%) \\
			\hline%\hline	
			\multirow{5}{*}{ResNet-34} &FPGM*~\cite{he2019filter}  & 73.92 & 91.62 & 1.38 & 0.49 & 41.1 \\
		&	BWCP* (Ours)  & 73.72 & 91.64 & \textbf{0.54} & \textbf{0.36} & \textbf{41.3}\\
			\cdashline{2-7}
			&SFP~\cite{he2018soft} & 73.92 & 91.62 & 2.09 & 1.29 & 41.1 \\ 
			&FPGM~\cite{he2019filter}   & 73.92 & 91.62  & 2.13 & 0.92 &41.1 \\
		&	BWCP (Ours)  & 73.72 & 91.64 & \textbf{1.27} & \textbf{0.86} & \textbf{41.3} \\
			\hline\hline
			\multirow{11}{*}{ResNet-50}&FPGM*~\cite{he2019filter}  & 76.15 & 92.87 & 1.32 & {0.55} & \textbf{53.5} \\
			&ThiNet* \cite{luo2017thinet}
			  & 72.88 & 91.14 & {0.84} & {0.47} & {36.8}\\
			&BWCP* (Ours)
			  & 76.20 & 93.15 & \textbf{0.52} & \textbf{0.41} & {51.8}\\
			\cdashline{2-7}
			&SSS~\cite{huang2018data}  & 76.12 & 92.86 & 4.30 & 2.07 & 43.0 \\
			&DSA~\cite{ning2020dsa} & -- & -- & 0.92 & \textbf{0.41} & 40.0 \\
			&HRank*~\cite{lin2020hrank}  & 76.15 & 92.87 & 1.17 & {0.64} & \textbf{43.7} \\
			&BWCP (Ours)  & 76.20 & 93.15 & \textbf{0.70} & 0.46 & {43.1}\\ 
			%\cdashline{2-8}
			&FPGM~\cite{he2019filter}   &76.15 & 92.87  & 2.02 & 0.93 & 53.5 \\
			&SCP \cite{kang2020operation}  & 75.89 & 92.98 & {1.69} & 0.98 & \textbf{54.3} \\
			&DSA~\cite{ning2020dsa}  & -- & -- & 1.33 & 0.80 & 50.0 \\
			
			&BWCP (Ours)  & 76.20 & 93.15 & \textbf{1.20} & \textbf{0.63 }& {51.8}\\
			\bottomrule		
		\end{tabular}
	}
	\label{table:ImageNet_table}
\vspace{-0.1in}
\end{table*}

\textbf{Inference Acceleration.} We analyze the realistic hardware acceleration in terms of GPU and CPU running time during inference. The CPU type is Intel Xeon CPU E5-2682 v4, and the GPU
is NVIDIA GTX1080TI. 
%The implementation is based on PyTorch 1.3. 
We evaluate the inference time using Resnet-50 with a mini-batch of 32 (1) on GPU (CPU). GPU inference batch size is larger than CPU
to emphasize our method's acceleration on the highly parallel platform as a structured pruning method. We see that BWCP has $29.2$\% inference time reduction on GPU, from $48.7$ms for base ResNet-50 to $34.5$ms for pruned ResNet-50, and $21.2$\% inference time reduction on CPU, from $127.1$ms for base ResNet-50 to $100.2$ms for pruned ResNet-50.
\vspace{-0.075in}
\subsection{Ablation Study}\label{sec:ablation}
\begin{figure}
\begin{center}
\includegraphics[scale=0.33]{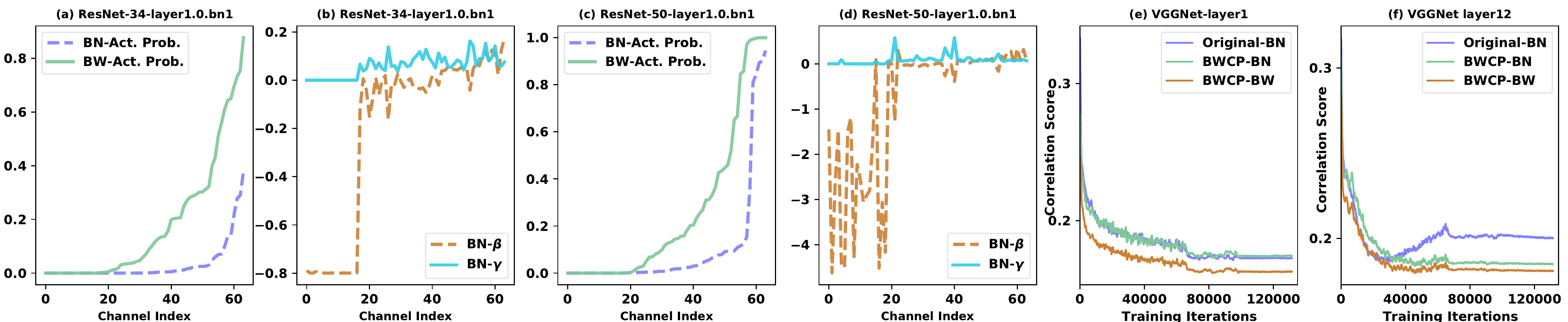}
\end{center}
  \caption{((\textbf{a}) \& (\textbf{b})) and ((\textbf{c}) \& (\textbf{d})) show the effect of BWCP on activation probability with trained ResNet-34 and ResNet-50 on ImageNet, respectively. 
  %We calculate the activation probability through Proposition \ref{remark:2}. We see that 
  The proposed batch whitening (BW) can increase the activation probability of useful channels when $|\gamma_c|>0$ while keeping the unimportant channels unchanged when when $\beta_c\leq 0$ and $\gamma_c \rightarrow 0$.
  (\textbf{e}) \& (\textbf{f}) show the correlation score for the output response maps in shallow  and deeper BWCP modules during the whole training period. 
%\textbf{(c)} shows the curves of correlation score at different depths. Results are obtained by applying VGGNet as backbone. 
BWCP has lower
correlation score among feature channels than original BN baseline.}
 \vspace{-0.25in}
\label{fig:analysis-6}
\end{figure}

\vspace{-0.075in}
\textbf{Effect of BWCP on activation probability.} From the analysis in Sec. \ref{sec:BW-analysis}, we have shown that BWCP can increase the activation probability of useful channels while keeping the activation probability of unimportant channels unchanged through BW technique. Here we demonstrate this using Resnet-34 and Resnet-50 trained on ImageNet dataset. We calculate the activation probability of channels of BN and BW layer. It can be seen from Fig.\ref{fig:analysis-6} (a-d) that (1) BW increases the activation probability of important channels when $|\gamma_c|>0$; (2) BW keeps the  the activation probability of unimportant channels unchanged when $\beta_c\leq 0$ and $\gamma_c \rightarrow 0$. Therefore, BW indeed works by making useful channels more important and unnecessary channels less important, respectively. In this way, BWCP can identify unimportant channels reliably.

\textbf{Effect of BW, Gumbel-Softmax (GS) and sparse Regularization (Reg).} The proposed BWCP consists of three components including BW module (\ie Eqn. (\ref{eq:BW-gb})) and Soft Sampling module with Gumbel-Softmax (\ie Eqn. (\ref{eq:soft-gate})) and a spare regularization (\ie Eqn. (\ref{eq:sparse_loss})). Here we investigate the effect of each component. To this end, five variants of BWCP are considered: (1) only BW module is used; (2) only sparse regularization is imposed; (3)BWCP w/o BW module; (4) BWCP w/o sparse regularization; and (5) BWCP with Gumbel-Softmax replaced by Straight Through Estimator (STE) \cite{BengioLC13}. For case (5), we select channels by hard $0$-$1$ mask generated with $m_c=\mathrm{sign}( P(\hat{{X}}_c>0)-0.5)$ \footnote{$y=\mathrm{sign}(x)=1$ if $x\geq 0$ and $0$ if $x<0$.}. The gradient is back-propagated through STE. From results on Table \ref{tab:BWCP-analy}, 
\begin{table}[t]
 \begin{minipage}[!t]{0.48\textwidth}
  \centering
     \makeatletter\def\@captype{table}\makeatother\caption{Effect of BW, Gumbel-Softmax (GS), and sparse Regularization in BWCP. The results are obtained by training ResNet-56 on CIFAR-10 dataset. `BL' denotes baseline model. }
       \scalebox{0.65}{
		\begin{tabular}{c c c c c c c}
			\toprule
			Cases& BW &GS &Reg &  Acc. (\%) & Model Size $\downarrow$  & FLOPs $\downarrow$ \\
			\hline	
			BL &\xmark & \xmark & \xmark &93.64 & - & - \\
			(1)&\cmark & \xmark & \xmark &\textbf{94.12}  & - & -  \\
			(2)&\xmark & \xmark & \cmark &93.46  & - & -  \\
			(3)&\xmark & \cmark & \cmark &92.84  & \textbf{46.37} & {51.16}  \\
			(4)&\cmark & \cmark & \xmark &94.10 & 7.78 & 6.25  \\
			(5)&\cmark & \xmark & \cmark &92.70  & 45.22 & \textbf{51.80}  \\
			BWCP&\cmark & \cmark & \cmark &93.37  & 44.42 & 50.35  \\

			\bottomrule			
		\end{tabular}
	}
	\label{tab:BWCP-analy}
  \end{minipage}
  \ \ \
  \begin{minipage}[!t]{0.48\textwidth}
   \centering
        \makeatletter\def\@captype{table}\makeatother\caption{	Effect of regularization strength $\lambda_1$ and $\lambda_2$ with magnitude $1e-4$ for the sparsity loss in Eqn.(\ref{eq:sparse_loss}). The results are obtained using VGG-16 on CIFAR-100 dataset.}
         \scalebox{0.78}{
		\begin{tabular}{c c c c c}
			\toprule
			 $\lambda_1$ & $\lambda_2 $ & Acc. (\%) & Acc. Drop & FLOPs $\downarrow$ (\%) \\
			\hline	
			  1.2 & 0.6 & 73.85 & -0.34 & 33.53 \\
			  1.2 & 1.2 & 73.66 &-0.15 & 35.92 \\
			  1.2 & 2.4 & 73.33 & 0.18 & 54.19 \\
			  0.6 & 1.2 & 74.27 & -0.76 & 30.67 \\
			  2.4 & 1.2 & 71.73 & 1.78 & 60.75 \\
			\bottomrule			
		\end{tabular}
	}
	\label{table:sensitivity_for_sparse}
	%\vspace{-0.2in}
   \end{minipage}
 \vspace{-0.1in}
\end{table}
we can make the following conclusions: \textbf{(a)} BW improves the recognition performance, implying that it can enhance the representation of channels; \textbf{(b)} sparse regularization on $\gamma$ and $\beta$ slight harm the classification accuracy of original model but it encourages channels to be sparse as also shown in Proposition \ref{remark:3}; (3) BWCP with Gumbel-Softmax achieves higher accuracy than STE, showing that a soft sampling technique is better than the deterministic ones as reported in \cite{jang2017categorical}.

\textbf{Impact of regularization strength $\lambda_1$ and $\lambda_2$.}
We analyze the effect of regularization strength $\lambda_1$ and $\lambda_2$ for sparsity loss on CIFAR-100.  The trade-off between accuracy and FLOPs reduction is investigated using VGG-16.
Table~\ref{table:sensitivity_for_sparse} illustrates that the network becomes more compact as $\lambda_1$ and $\lambda_2$ increase,  implying that both terms in Eqn.(\ref{eq:sparse_loss}) can make channel features sparse. Moreover, the flops metric is more sensitive to the regularization on $\beta$, which validates our  analysis in Sec.\ref{sec:BW-analysis}). Besides, we should search for proper values for $\lambda_1$ and $\lambda_2$ to trade off between accuracy and FLOPs reduction, which is a drawback for our method.

\textbf{Effect of the number of BW.} 
Here the effect of the number of BW modules of BWCP is investigated trained on CIFAR-10 using Resnet-56 consisting of a series of bottleneck structures. Note that there are three BN layers in each bottleneck. We study four variants of BWCP: 
%(a) no BW is employed, which indicates that the number of BW module is zero; 
%
\begin{wraptable}{r}{7.0cm}
	\centering
	\vspace{-0.2in}
	\caption{
		Effect of the number of BW modules on CIFAR-10 dataset trained with ResNet-56. `\# BW' indicates the number of BW. 
		%We study four variants of the proposed BWCP where $\{18,36,54\}$ BW layers are plugged into ResNet-56, respectively. We see that 
		More BW modules in the network would lead to a lower recognition accuracy drop with comparable computation consumption.
	}
	%\vspace{-0.1in}
	\scalebox{0.70}{
		\begin{tabular}{c c c c c}
			\toprule
			 \# BW  &  Acc. (\%) & Acc. Drop & Model Size $\downarrow$ (\%)  & FLOPs $\downarrow$ (\%) \\
			\hline	
			  %0 &  92.84 & 0.80 & 46.37 & 51.16 \\
			  18 & 93.01  & 0.63 & 44.70 & 50.77 \\
			  36 & 93.14  & 0.50 & 45.29 & 50.45 \\
			  54 & 93.37 & 0.27 & 44.42 & 50.35 \\
			\bottomrule			
		\end{tabular}
	}
	\vspace{-0.2in}
	\label{tab:num-BW}
\end{wraptable}
(a) we use BW to modify the last BN in each bottleneck module; hence there are a total of 18 BW layers in Resnet-56; (b) the last two BN layers are modified by our BW technique (36 BW layers) (c) All BN layers in bottlenecks are replaced by BW  (54 BW layers), which is our proposed method. The results are reported in Table \ref{tab:num-BW}. We can see that BWCP achieves the best top-1 accuracy when BW acts on all BN layers, given the comparable FLOPs and model size. This indicates that the proposed BWCP more benefits from more BW layers in the network.
%
\iffalse
\begin{wraptable}{r}{6cm}
	\centering
	\caption{
		Effect of regularization strength $\lambda_1$ and $\lambda_2$ with magnitude $1e-4$ for the sparsity loss in Eqn.(\ref{eq:sparse_loss}). The results are obtained using VGG-16 on CIFAR-100 dataset. 
	} %($\mathcal{L}_{\text{sparse}}$)
	%\vspace{-0.1in}
	\scalebox{0.80}{
		\begin{tabular}{c c c c c}
			\toprule
			 $\lambda_1$ & $\lambda_2 $ & Acc. (\%) & Acc. Drop & FLOPs $\downarrow$ (\%) \\
			\hline	
			  1.2 & 0.6 & 73.85 & -0.34 & 33.53 \\
			  1.2 & 1.2 & 73.66 &-0.15 & 35.92 \\
			  1.2 & 2.4 & 73.33 & 0.18 & 54.19 \\
			  0.6 & 1.2 & 74.27 & -0.76 & 30.67 \\
			  2.4 & 1.2 & 71.73 & 1.78 & 60.75 \\
			\bottomrule			
		\end{tabular}
	}
	\label{table:sensitivity_for_sparse}
	%\vspace{-0.2in}
\end{wraptable}
%
\begin{wrapfigure}{r}{5.5cm}
\centering
\includegraphics[scale=0.33]{./Corr.pdf}
\vspace{-0.1in}
  \caption{\textbf{(a)} \& \textbf{(b)} show the correlation score for the output response maps in shallow (Left) and deeper (Middle) BW modules
during the whole training period. 
%\textbf{(c)} shows the curves of correlation score at different depths. Results are obtained by applying VGGNet as backbone. 
BWCP has lower
correlation score among feature channels than original BN baseline.}
\label{fig:Corr-BW}
\vspace{-0.2in}
\end{wrapfigure}
\fi

\textbf{BWCP learns representative channel features.} It is worth noting that BWCP can whiten channel features after BN through BW as shwon in Eqn.(\ref{eq:BW-gb}).
%when $\mathbf{\Sigma}$ is diagonal in Eqn.(\ref{eq:BW-gb}). To see this, we rewrite Eqn.(\ref{eq:BW-gb}) as $\hat{\mathbf{x}}_{nij} = (\Sigmainv\mathbf{\gamma}) \odot \bar{\mathbf{x}}_{nij} + \Sigmainv\mathbf{\beta}=\Sigmainv (\mathbf{\gamma}\odot \bar{\mathbf{x}}_{nij}) + \Sigmainv\mathbf{\beta}=\Sigmainv\mathbf{\tilde{x}}_{nij}$, which shows that whitening operator can also perform on channel features. 
Therfore, BW  can learn diverse channel features by reducing the correlations among channels\cite{yang2019cross}. 
We investigate this using VGGNet-16 with BN and the proposed BWCP trained on CIFAR-10. The correlation score can be calculated by taking the average over the absolute value of the correlation matrix of channel features. A larger value indicates that there is redundancy in the encoded features. We plot the correlation score among
channels at different depths of the network. As shown in Fig.\ref{fig:analysis-6} (e \& f), we can see that channel features after BW block have significantly smaller correlations, implying that BWCP learns representative channel features. This also accounts for the effectiveness of the proposed scheme.
\vspace{-0.1in}
\section{Discussion and Conclusion}
\vspace{-0.1in}
In this paper, we presented an effective and efficient pruning technique, termed Batch Whitening Channel Pruning (BWCP). We show
that BWCP increases the activation probability of useful channels while keeping unimportant channels unchanged, making it appealing for pursuing a compact model. 
Particularly, BWCP can be  easily applied to prune various CNN architecture by modifying the batch normalization layer. However, the proposed BWCP needs to exhaustively search the strength of sparse regularization to achieve different levels of FLOPs reduction. With probabilistic formulation in BWCP, the expected FLOPs can be modeled. The multiplier method can be used to encourage the model to attain target FLOPs. For future work, an advanced Pareto optimization algorithm can be designed to tackle such multi-objective joint minimization.  We hope that the analyses of BWCP could bring a
new perspective for future work in channel pruning.

% In the unusual situation where you want a paper to appear in the
% references without citing it in the main text, use \nocite
%\nocite{langley00}

\begin{ack}
We thank Minsoo Kang, Shitao Tang and Ting Zhang for their helpful discussions. We also thank anonymous reviewers from venues where we have submitted this work for their instructive feedback.
\end{ack}

\bibliography{main}

\begin{thebibliography}{10}

\bibitem{he2016deep}
Kaiming {He}, Xiangyu {Zhang}, Shaoqing {Ren}, and Jian {Sun}.
\newblock Deep residual learning for image recognition.
\newblock In {\em 2016 IEEE Conference on Computer Vision and Pattern
  Recognition (CVPR)}, pages 770--778, 2016.

\bibitem{he2016identity}
Kaiming {He}, Xiangyu {Zhang}, Shaoqing {Ren}, and Jian {Sun}.
\newblock Identity mappings in deep residual networks.
\newblock In {\em European Conference on Computer Vision}, pages 630--645,
  2016.

\bibitem{ren2017faster}
Shaoqing {Ren}, Kaiming {He}, Ross {Girshick}, and Jian {Sun}.
\newblock Faster r-cnn: Towards real-time object detection with region proposal
  networks.
\newblock {\em IEEE Transactions on Pattern Analysis and Machine Intelligence},
  39(6):1137--1149, 2017.

\bibitem{he2020mask}
Kaiming {He}, Georgia {Gkioxari}, Piotr {Dollar}, and Ross {Girshick}.
\newblock Mask r-cnn.
\newblock {\em IEEE Transactions on Pattern Analysis and Machine Intelligence},
  42(2):386--397, 2020.

\bibitem{chen2018deeplab}
Liang-Chieh {Chen}, George {Papandreou}, Iasonas {Kokkinos}, Kevin {Murphy},
  and Alan~L. {Yuille}.
\newblock Deeplab: Semantic image segmentation with deep convolutional nets,
  atrous convolution, and fully connected crfs.
\newblock {\em IEEE Transactions on Pattern Analysis and Machine Intelligence},
  40(4):834--848, 2018.

\bibitem{shelhamer2017fully}
Evan {Shelhamer}, Jonathan {Long}, and Trevor {Darrell}.
\newblock Fully convolutional networks for semantic segmentation.
\newblock {\em IEEE Transactions on Pattern Analysis and Machine Intelligence},
  39(4):640--651, 2017.

\bibitem{han2018bandwidth}
Song {Han} and William~J. {Dally}.
\newblock Bandwidth-efficient deep learning.
\newblock In {\em Proceedings of the 55th Annual Design Automation Conference
  on}, page 147, 2018.

\bibitem{deng2020model}
Lei {Deng}, Guoqi {Li}, Song {Han}, Luping {Shi}, and Yuan {Xie}.
\newblock Model compression and hardware acceleration for neural networks: A
  comprehensive survey.
\newblock {\em Proceedings of the IEEE}, 108(4):485--532, 2020.

\bibitem{zhu2018to}
Michael~H. {Zhu} and Suyog {Gupta}.
\newblock To prune, or not to prune: Exploring the efficacy of pruning for
  model compression.
\newblock In {\em ICLR 2018 : International Conference on Learning
  Representations 2018}, 2018.

\bibitem{sun2017ensemble}
Shizhao {Sun}, Wei {Chen}, Jiang {Bian}, Xiaoguang {Liu}, and Tie-Yan {Liu}.
\newblock Ensemble-compression: A new method for parallel training of deep
  neural networks.
\newblock In {\em Joint European Conference on Machine Learning and Knowledge
  Discovery in Databases}, pages 187--202, 2017.

\bibitem{han2016deep}
Song {Han}, Huizi {Mao}, and William~J. {Dally}.
\newblock Deep compression: Compressing deep neural networks with pruning,
  trained quantization and huffman coding.
\newblock In {\em ICLR 2016 : International Conference on Learning
  Representations 2016}, 2016.

\bibitem{wen2016learning}
Wei {Wen}, Chunpeng {Wu}, Yandan {Wang}, Yiran {Chen}, and Hai {Li}.
\newblock Learning structured sparsity in deep neural networks.
\newblock In {\em Proceedings of the 30th International Conference on Neural
  Information Processing Systems}, pages 2074--2082, 2016.

\bibitem{han2015learning}
Song Han, Jeff Pool, John Tran, and William Dally.
\newblock Learning both weights and connections for efficient neural network.
\newblock In {\em Advances in neural information processing systems}, pages
  1135--1143, 2015.

\bibitem{li2016pruning}
Hao Li, Asim Kadav, Igor Durdanovic, Hanan Samet, and Hans~Peter Graf.
\newblock Pruning filters for efficient convnets.
\newblock {\em ICLR 2017 : International Conference on Learning Representations
  2018}, 2016.

\bibitem{guo2016dynamic}
Yiwen Guo, Anbang Yao, and Yurong Chen.
\newblock Dynamic network surgery for efficient dnns.
\newblock In {\em Advances in neural information processing systems}, pages
  1379--1387, 2016.

\bibitem{liu2018rethinking}
Zhuang Liu, Mingjie Sun, Tinghui Zhou, Gao Huang, and Trevor Darrell.
\newblock Rethinking the value of network pruning.
\newblock {\em arXiv preprint arXiv:1810.05270}, 2018.

\bibitem{liu2017learning}
Zhuang Liu, Jianguo Li, Zhiqiang Shen, Gao Huang, Shoumeng Yan, and Changshui
  Zhang.
\newblock Learning efficient convolutional networks through network slimming.
\newblock In {\em Proceedings of the IEEE International Conference on Computer
  Vision}, pages 2736--2744, 2017.

\bibitem{lin2020hrank}
Mingbao Lin, Rongrong Ji, Yan Wang, Yichen Zhang, Baochang Zhang, Yonghong
  Tian, and Ling Shao.
\newblock Hrank: Filter pruning using high-rank feature map.
\newblock In {\em Proceedings of the IEEE/CVF Conference on Computer Vision and
  Pattern Recognition}, pages 1529--1538, 2020.

\bibitem{frankle2018lottery}
Jonathan Frankle and Michael Carbin.
\newblock The lottery ticket hypothesis: Finding sparse, trainable neural
  networks.
\newblock {\em arXiv preprint arXiv:1803.03635}, 2018.

\bibitem{yang2019deephoyer}
Huanrui Yang, Wei Wen, and Hai Li.
\newblock Deephoyer: Learning sparser neural network with differentiable
  scale-invariant sparsity measures.
\newblock {\em arXiv preprint arXiv:1908.09979}, 2019.

\bibitem{luo2017thinet}
Jian-Hao Luo, Jianxin Wu, and Weiyao Lin.
\newblock Thinet: A filter level pruning method for deep neural network
  compression.
\newblock In {\em Proceedings of the IEEE international conference on computer
  vision}, pages 5058--5066, 2017.

\bibitem{he2019filter}
Yang He, Ping Liu, Ziwei Wang, Zhilan Hu, and Yi~Yang.
\newblock Filter pruning via geometric median for deep convolutional neural
  networks acceleration.
\newblock In {\em Proceedings of the IEEE Conference on Computer Vision and
  Pattern Recognition}, pages 4340--4349, 2019.

\bibitem{he2018soft}
Yang He, Guoliang Kang, Xuanyi Dong, Yanwei Fu, and Yi~Yang.
\newblock Soft filter pruning for accelerating deep convolutional neural
  networks.
\newblock {\em arXiv preprint arXiv:1808.06866}, 2018.

\bibitem{ioffe2015batch}
Sergey Ioffe and Christian Szegedy.
\newblock Batch normalization: Accelerating deep network training by reducing
  internal covariate shift.
\newblock {\em arXiv preprint arXiv:1502.03167}, 2015.

\bibitem{arpit2016normalization}
Devansh Arpit, Yingbo Zhou, Bhargava~U Kota, and Venu Govindaraju.
\newblock Normalization propagation: A parametric technique for removing
  internal covariate shift in deep networks.
\newblock {\em International Conference in Machine Learning}, 2016.

\bibitem{lecun1990optimal}
Yann LeCun, John~S Denker, and Sara~A Solla.
\newblock {Optimal Brain Damage}.
\newblock In {\em NIPS}, 1990.

\bibitem{louizos2017learning}
Christos Louizos, Max Welling, and Diederik~P Kingma.
\newblock Learning sparse neural networks through $ l\_0 $ regularization.
\newblock {\em International Conference on Learning Representation}, 2017.

\bibitem{li2017}
Hao Li, Asim Kadav, Igor Durdanovic, Hanan Samet, and Hans~Peter Graf.
\newblock {Pruning Filters for Efficient ConvNets}.
\newblock In {\em ICLR}, 2017.

\bibitem{zhao2019variational}
Chenglong Zhao, Bingbing Ni, Jian Zhang, Qiwei Zhao, Wenjun Zhang, and Qi~Tian.
\newblock Variational convolutional neural network pruning.
\newblock In {\em Proceedings of the IEEE Conference on Computer Vision and
  Pattern Recognition}, pages 2780--2789, 2019.

\bibitem{kang2020operation}
Minsoo Kang and Bohyung Han.
\newblock Operation-aware soft channel pruning using differentiable masks.
\newblock {\em arXiv preprint arXiv:2007.03938}, 2020.

\bibitem{perez2018film}
Ethan Perez, Florian Strub, Harm De~Vries, Vincent Dumoulin, and Aaron
  Courville.
\newblock Film: Visual reasoning with a general conditioning layer.
\newblock In {\em Proceedings of the AAAI Conference on Artificial
  Intelligence}, volume~32, 2018.

\bibitem{wang2020deep}
Yikai Wang, Wenbing Huang, Fuchun Sun, Tingyang Xu, Yu~Rong, and Junzhou Huang.
\newblock Deep multimodal fusion by channel exchanging.
\newblock {\em Advances in Neural Information Processing Systems}, 33, 2020.

\bibitem{bini2005algorithms}
Dario~A Bini, Nicholas~J Higham, and Beatrice Meini.
\newblock Algorithms for the matrix pth root.
\newblock {\em Numerical Algorithms}, 39(4):349--378, 2005.

\bibitem{huang2019iterative}
Lei Huang, Yi~Zhou, Fan Zhu, Li~Liu, and Ling Shao.
\newblock Iterative normalization: Beyond standardization towards efficient
  whitening.
\newblock In {\em Proceedings of the IEEE Conference on Computer Vision and
  Pattern Recognition}, pages 4874--4883, 2019.

\bibitem{jang2017categorical}
Eric Jang, Shixiang Gu, and Ben Poole.
\newblock Categorical reparameterization with gumbel-softmax.
\newblock 2017.

\bibitem{huang2017densely}
Gao Huang, Zhuang Liu, Laurens Van Der~Maaten, and Kilian~Q Weinberger.
\newblock Densely connected convolutional networks.
\newblock In {\em Proceedings of the IEEE conference on computer vision and
  pattern recognition}, pages 4700--4708, 2017.

\bibitem{zhuang2018discrimination}
Zhuangwei Zhuang, Mingkui Tan, Bohan Zhuang, Jing Liu, Yong Guo, Qingyao Wu,
  Junzhou Huang, and Jinhui Zhu.
\newblock Discrimination-aware channel pruning for deep neural networks.
\newblock {\em arXiv preprint arXiv:1810.11809}, 2018.

\bibitem{he2018amc}
Yihui He, Ji~Lin, Zhijian Liu, Hanrui Wang, Li-Jia Li, and Song Han.
\newblock Amc: Automl for model compression and acceleration on mobile devices.
\newblock In {\em Proceedings of the European Conference on Computer Vision
  (ECCV)}, pages 784--800, 2018.

\bibitem{krizhevsky2009learning}
A~Krizhevsky.
\newblock Learning multiple layers of features from tiny images.
\newblock {\em Technical report}, 2009.

\bibitem{russakovsky2015imagenet}
Olga Russakovsky, Jia Deng, Hao Su, Jonathan Krause, Sanjeev Satheesh, Sean Ma,
  Zhiheng Huang, Andrej Karpathy, Aditya Khosla, Michael Bernstein, et~al.
\newblock Imagenet large scale visual recognition challenge.
\newblock {\em International Journal of Computer Vision}, 115(3):211--252,
  2015.

\bibitem{simonyan2014very}
Karen Simonyan and Andrew Zisserman.
\newblock Very deep convolutional networks for large-scale image recognition.
\newblock {\em arXiv preprint arXiv:1409.1556}, 2014.

\bibitem{ye2020good}
Mao Ye, Chengyue Gong, Lizhen Nie, Denny Zhou, Adam Klivans, and Qiang Liu.
\newblock Good subnetworks provably exist: Pruning via greedy forward
  selection.
\newblock In {\em International Conference on Machine Learning}, pages
  10820--10830. PMLR, 2020.

\bibitem{huang2018data}
Zehao Huang and Naiyan Wang.
\newblock {Data-Driven Sparse Structure Selection for Deep Neural Networks}.
\newblock In {\em ECCV}, 2018.

\bibitem{ning2020dsa}
Xuefei Ning, Tianchen Zhao, Wenshuo Li, Peng Lei, Yu~Wang, and Huazhong Yang.
\newblock Dsa: More efficient budgeted pruning via differentiable sparsity
  allocation.
\newblock {\em arXiv preprint arXiv:2004.02164}, 2020.

\bibitem{BengioLC13}
Yoshua Bengio, Nicholas L{\'{e}}onard, and Aaron~C. Courville.
\newblock Estimating or propagating gradients through stochastic neurons for
  conditional computation.
\newblock {\em CoRR}, abs/1308.3432, 2013.

\bibitem{yang2019cross}
Jianwei Yang, Zhile Ren, Chuang Gan, Hongyuan Zhu, and Devi Parikh.
\newblock Cross-channel communication networks.
\newblock In {\em Advances in Neural Information Processing Systems}, pages
  1295--1304, 2019.

\end{thebibliography}
\bibliographystyle{unsrt}

%%%%%%%%%%%%%%%%%%%%%%%%%%%%%%%%%%%%%%%%%%%%%%%%%%%%%%%%%%%%%%%%%%%%%%%%%%%%%%%
%%%%%%%%%%%%%%%%%%%%%%%%%%%%%%%%%%%%%%%%%%%%%%%%%%%%%%%%%%%%%%%%%%%%%%%%%%%%%%%
% DELETE THIS PART. DO NOT PLACE CONTENT AFTER THE REFERENCES!
%%%%%%%%%%%%%%%%%%%%%%%%%%%%%%%%%%%%%%%%%%%%%%%%%%%%%%%%%%%%%%%%%%%%%%%%%%%%%%%
%%%%%%%%%%%%%%%%%%%%%%%%%%%%%%%%%%%%%%%%%%%%%%%%%%%%%%%%%%%%%%%%%%%%%%%%%%%%%%%
\newpage
\setcounter{section}{0}
\renewcommand\thesection{\Alph{section}}
\title{Appendix of BWCP: Probabilistic Learning-to-Prune Channels for ConvNets via Batch Whitening}
\maketitle

The appendix provides more details about approach and experiments of our proposed batch whitening channel pruning (BWCP) framework.

%%%%%%%%% BODY TEXT
\section{More Details about Approach}

%-------------------------------------------------------------------------
\subsection{Calculation of Covariance Matrix $\mathbf{\Sigma}$}\label{sec:A.1}
  By Eqn.(\ref{eq:conv-norm-relu}) in main text, the output of BN is $\tilde{x}_{ncij}=\gamma_c\bar{x}_{ncij}+\beta_c$. Hence, we have $\mathbb{E}[\tilde{\mathbf{x}}_c] =  \frac{1}{NHW}\sum^{N,H,W}_{n,i,j}(\gamma_c\bar{x}_{ncij}+\beta_c) = \beta_c$. Then the entry in $c$-th row and $d$-th column of covariance matrix $\Sigma$ of $\tilde{\mathbf{x}}_c$ is calculated as follows:
\begin{equation}
\Sigma_{cd}=\frac{1}{NHW}\sum^{N,H,W}_{n,i,j}(\gamma_c\bar{x}_{ncij}+\beta_c - \mathbb{E}[\tilde{\mathbf{x}}_c])(\gamma_d\bar{x}_{ndij}+\beta_d-\mathbb{E}[\tilde{\mathbf{x}}_d])=\gamma_c\gamma_d\rho_{cd}
\end{equation}
where $\rho_{cd}$ is the element in c-th row and j-th column of correlation matrix of $\bar{x}$. Hence, we have $\rho_{cd}\in [-1,1]$. Furthermore, we can write $\Sigma$ into the vector form: $\mathbf{\Sigma}=\boldsymbol{\gamma}\boldsymbol{\gamma}\tran \odot \frac{1}{NHW}\sum^{N,H,W}_{n,i,j}\bar{\mathbf{x}}_{nij}\bar{\mathbf{x}} _{nij}\tran=\boldsymbol{\gamma}\boldsymbol{\gamma}\tran \odot \boldsymbol{\rho}$.
%-------------------------------------------------------------------------
\subsection{Proof of Proposition \ref{remark:1}}\label{sec:A.2}
For (1),  we notice that we can define ${\gamma}_c = -{\gamma}_c$ and $\bar{X}_c = -\bar{X}_c \sim \mathcal{N}(0,1)$ if $\gamma_c <0$. Hence, we can assume ${\gamma}_c >0$ without loss of generality. Then, we have
\begin{equation}
\begin{split}
\vspace{-0.1in}
P(Y_c>0)=P({\tilde{{X}}}_c>0&)= P({\bar{{X}}}_c>-\frac{\beta_c}{\gamma_c})\\
&= \int_{-\frac{\beta_c}{\gamma_c}}^{+\infty}\frac{1}{\sqrt{2\pi}}\mathrm{exp}^{-\frac{t^2}{2}}dt\\
&= \int_{-\frac{\beta_c}{\gamma_c}}^{0}\frac{1}{\sqrt{2\pi}}\mathrm{exp}^{-\frac{t^2}{2}}dt + \int_{0}^{+\infty}\frac{1}{\sqrt{2\pi}}\mathrm{exp}^{-\frac{t^2}{2}}dt\\
&= \int^{\frac{\beta_c}{\gamma_c}}_{0}\frac{1}{\sqrt{2\pi}}\mathrm{exp}^{-\frac{t^2}{2}}dt + \int_{0}^{+\infty}\frac{1}{\sqrt{2\pi}}\mathrm{exp}^{-\frac{t^2}{2}}dt\\
&= \frac{\mathrm{Erf}(\frac{\beta_c}{\sqrt{2}\gamma_c})+1}{2}
\end{split}
\vspace{-0.1in}
\end{equation}
When $\gamma_c <0$, we can set ${\gamma}_c = -{\gamma}_c$. Hence, we arrive at
\begin{equation}
P(Y_c>0)=P({\tilde{{X}}}_c>0)=\frac{\mathrm{Erf}(\frac{\beta_c}{\sqrt{2}|\gamma_c|})+1}{2}
\end{equation}
For (2), let us denote $\bar{{X}}_c \sim \mathcal{N}(0,1)$,   and $\tilde{X}_c=\gamma_c \bar{X}_c +\beta_c$ and $Y_c=\max\{0,\tilde{X}_c\}$ where $Y_c$ represents a random variables corresponding to output feature $\mathbf{y}_c$ in Eqn.(1) in main text.
Firstly, it is easy to see that $P(\tilde{{X}}_c>0) = 0 \Leftrightarrow \mathbb{E}_{\bar{X}_c}[Y_c]=0$ and $\mathbb{E}_{\bar{X}_c}[Y_c^2]=0$. In the following we show that $ \mathbb{E}_{\bar{X}_c}[Y_c]=0$ and $\mathbb{E}_{\bar{X}_c}[Y_c^2]=0 \Leftrightarrow \beta_c \leq 0\, \mathrm{and}\, {\gamma_c=0}$.
 Similar with (1), we assume $\gamma_c>0$ without loss of generality.
 
 For the sufficiency, we have
\begin{equation}\label{eq:limgamma1}
\begin{split}
\mathbb{E}_{\bar{X}_c}[Y_c]&=\int_{-\infty}^{-\frac{\beta_c}{\gamma_c}}0\cdot \frac{1}{\sqrt{2\pi}}\mathrm{exp}^{-\frac{{\bar{x}_c}^2}{2}}d{\bar{x}_c}
+\int_{-\frac{\beta_c}{\gamma_c}}^{+\infty}(\gamma_c{\bar{x}_c}+\beta_c)\cdot \frac{1}{\sqrt{2\pi}}\mathrm{exp}^{-\frac{{\bar{x}_c}^2}{2}}d{\bar{x}_c},\\
&=\frac{\gamma_c\mathrm{exp}^{-\frac{\beta_c^2}{2\gamma_c^2}}}{\sqrt{2\pi}}+\frac{\beta_c}{2}(1+\mathrm{Erf}[\frac{\beta_c}{\sqrt{2}\gamma_c}]),
\end{split}
\end{equation}
where $\mathrm{Erf}[x]=\frac{2}{\sqrt{\pi}}\int_0^{x}\mathrm{exp}^{-t^2}dt$ is the error function. From Eqn.(\ref{eq:limgamma1}), we have
\begin{equation}\label{eq:limgamma2}
\begin{split}
\lim_{\gamma_c\rightarrow 0^+}\mathbb{E}_{\bar{X}_c}[{Y_c}]&=\lim_{\gamma_c\rightarrow 0^+}\frac{\gamma_c\mathrm{exp}^{-\frac{\beta_c^2}{2\gamma_c^2}}}{\sqrt{2\pi}}+\lim_{\gamma_c\rightarrow 0^+}\frac{\beta_c}{2}(1+\mathrm{Erf}[\frac{\beta_c}{\sqrt{2}\gamma_c}])=0
\end{split}
\end{equation}
In the same way, we can calculate
\begin{equation}\label{eq:limgamma3}
\begin{split}
\mathbb{E}_{{\bar{x}_c}}[Y_c^2]&=\int_{-\infty}^{-\frac{\beta_c}{\gamma_c}}0\cdot \frac{1}{\sqrt{2\pi}}\mathrm{exp}^{-\frac{{\bar{x}_c}^2}{2}}d{\bar{x}_c}+\int_{-\frac{\beta_c}{\gamma_c}}^{+\infty}(\gamma_c{\bar{x}_c}+\beta_c)^2\cdot \frac{1}{\sqrt{2\pi}}\mathrm{exp}^{-\frac{{\bar{x}_c}^2}{2}}d{\bar{x}_c},\\
&=\frac{\gamma_c\beta_c\mathrm{exp}^{-\frac{\beta_c^2}{2\gamma_c^2}}}{\sqrt{2\pi}}+\frac{\gamma_c^2+\beta_c^2}{2}(1+\mathrm{Erf}[\frac{\beta_c}{\sqrt{2}\gamma_c}]),
\end{split}
\end{equation}
From Eqn.(\ref{eq:limgamma3}), we have
\begin{equation}\label{eq:limgamma4}
\begin{split}
\lim_{\gamma_c\rightarrow 0^+}\mathbb{E}_{{\bar{x}_c}}[Y_c^2]&=\lim_{\gamma_c\rightarrow 0^+}\frac{\gamma_c\beta_c\mathrm{exp}^{-\frac{\beta_c^2}{2\gamma_c^2}}}{\sqrt{2\pi}}+\lim_{\gamma_c\rightarrow 0^+}\frac{\gamma_c^2+\beta_c^2}{2}(1+\mathrm{Erf}[\frac{\beta_c}{\sqrt{2}\gamma_c}])=0
\end{split}
\end{equation}

For necessity, we show that if $\mathbb{E}_{{\bar{x}_c}}[{Y_c}]=0$ and $\mathbb{E}_{{\bar{x}_c}}[{Y_c}^2]=0$, then $\gamma_c \rightarrow 0$ and $\beta_c\leq 0$. In essence, It can be acquired by solving Eqn Eqn.(\ref{eq:limgamma1}) and Eqn.(\ref{eq:limgamma3}). To be specific, $\beta_c*$Eqn.(.\ref{eq:limgamma1})$-$Eqn.(\ref{eq:limgamma3}) gives us $\gamma_c=0^+$. Substituting it into Eqn.(.\ref{eq:limgamma1}), we can obtain $\beta_c \leq 0$.
This completes the proof.

\subsection{Proof of Proposition \ref{remark:2}}\label{sec:A.3}

First, we can derive that $\hat{X}_c = \SigmainvN (\boldsymbol{\gamma}\odot \bar{{X}}+\boldsymbol{\beta}) = \SigmainvN (\boldsymbol{\gamma} \odot \bar{{X}})+\SigmainvN\boldsymbol{\beta} = (\SigmainvN\boldsymbol{\gamma}) \odot \bar{X}+\SigmainvN \boldsymbol{\beta}$. 

Hence, the newly defined scale and bias parameters are  $\hat{\boldsymbol{\gamma}} = \SigmainvN\boldsymbol{\gamma}$ and $\hat{\boldsymbol{\beta}} = \SigmainvN\boldsymbol{\beta}$.
When $T=1$, we have $\SigmainvN = \frac{1}{2}(3\mathbf{I}-\mathbf{\Sigma}_N)$ by Eqn.(\ref{eq:Newtoniter}) in main text. Hence we obtain,
\begin{equation}\label{eq:gamma2}
\begin{split}
    \hat{\boldsymbol{\gamma}}&=\frac{1}{2}(3\mathbf{I}-\mathbf{\Sigma}_N)\boldsymbol{\gamma}=\frac{1}{2}(3\mathbf{I}-\frac{\boldsymbol{\gamma}\boldsymbol{\gamma}\tran}{\left\|\boldsymbol{\gamma}\right\|_2^2} \odot\boldsymbol{\rho})\boldsymbol{\gamma}\\
     &=\frac{1}{2}(3\boldsymbol{\gamma}-\frac{1}{\left\|\boldsymbol{\gamma}\right\|_2^2}\left[\sum_{j=1}^C\gamma_1\gamma_j\rho_{1j}\gamma_j,\cdots,\sum_{j=1}^C\gamma_C\gamma_j\rho_{Cj}\gamma_j\right]^{\mathrm{T}})\\
     &=\frac{1}{2}\left[(3-\sum_{j=1}^C\frac{\gamma_j^2\rho_{1j}}{\left\|\boldsymbol{\gamma}\right\|_2^2})\gamma_1,\cdots,(3-\sum_{j=1}^C\frac{\gamma_j^2\rho_{Cj}}{\left\|\boldsymbol{\gamma}\right\|_2^2})\gamma_C\right]^{\mathrm{T}}
\end{split}
\end{equation}
Similarly, $\hat{\boldsymbol{\beta}}$ can be given by
\begin{equation}\label{eq:beta2}
\begin{split}
    \hat{\boldsymbol{\beta}}&=\frac{1}{2}(3\mathbf{I}-\mathbf{\Sigma})\boldsymbol{\beta}=\frac{1}{2}(3\mathbf{I}-\frac{\boldsymbol{\gamma}\boldsymbol{\gamma}\tran}{\left\|\boldsymbol{\gamma}\right\|_2^2} \odot\boldsymbol{\rho})\boldsymbol{\beta}\\
    &=\frac{1}{2}(3\boldsymbol{\beta}-\frac{1}{\left\|\boldsymbol{\gamma}\right\|_2^2}\left[\sum_{j=1}^C\gamma_1\gamma_j\rho_{1j}\beta_j,\cdots,\sum_{j=1}^C\gamma_C\gamma_j\rho_{Cj}\beta_j\right]^{\mathrm{T}})\\
     &=\frac{1}{2}\left[3\beta_1-(\sum_{j=1}^C\frac{\gamma_j\beta_j\rho_{1j}}{\left\|\boldsymbol{\gamma}\right\|_2^2})\gamma_1,\cdots,3\beta_C-(\sum_{j=1}^C\frac{\gamma_j\beta_j\rho_{Cj}}{\left\|\boldsymbol{\gamma}\right\|_2^2})\gamma_C\right]^{\mathrm{T}}
\end{split}
\end{equation}
Taking each component of vector Eqn.(\ref{eq:gamma2}-\ref{eq:beta2}) gives us the expression of $\hat{{\gamma}_c}$ and $\hat{{\beta}_c}$ in Proposition \ref{remark:2}.

\subsection{Proof of Proposition \ref{remark:3}}\label{sec:A.4}

\begin{figure*}
\begin{center}
\includegraphics[scale=0.3]{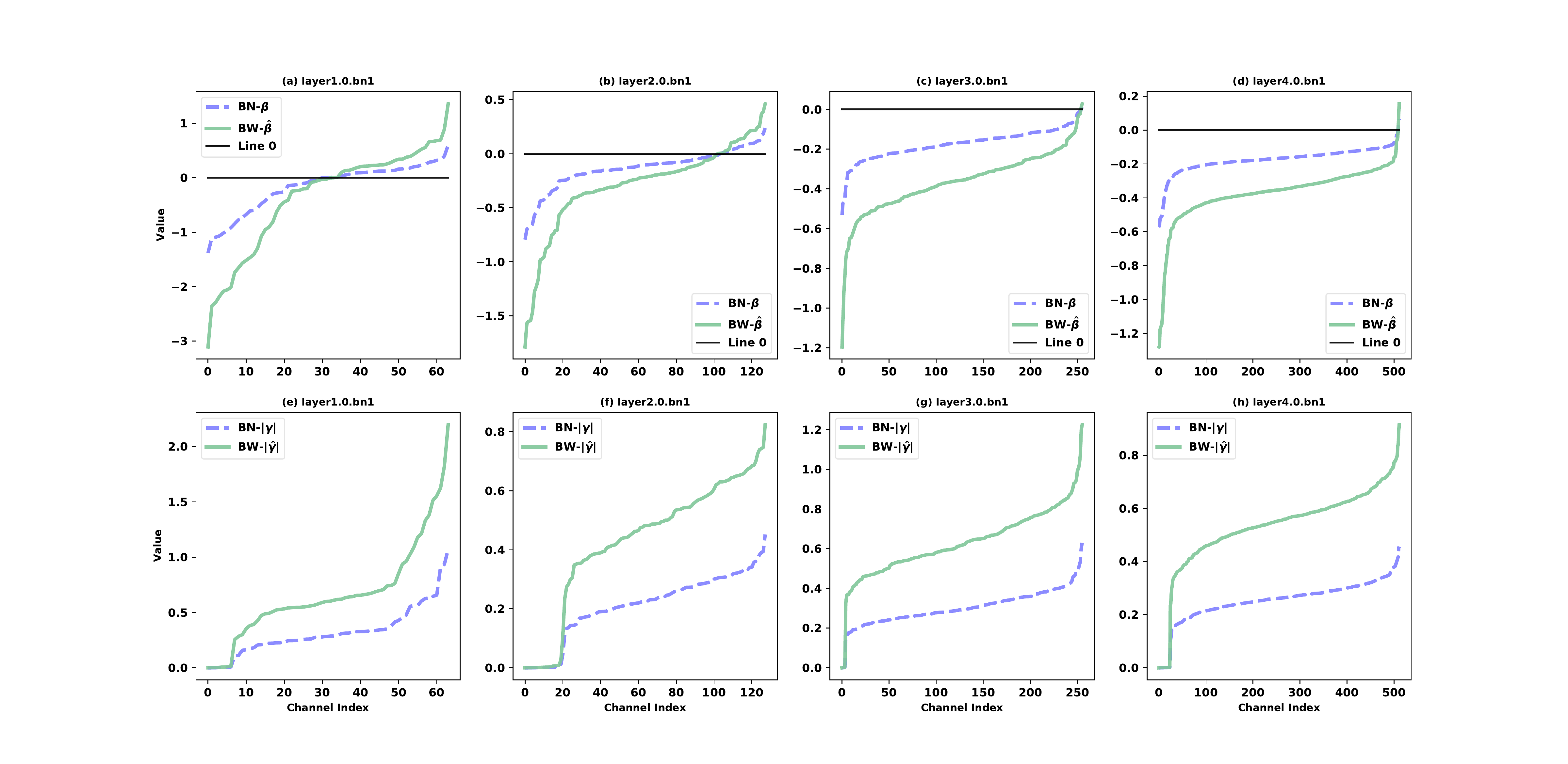}
\end{center}
\vspace{-0.1in}
  \caption{Experimental observation of how our proposed BWCP changes the values of $\gamma_c$ and $\beta_c$ in BN layers through the proposed BW technique. Results are obtained by tranining ResNet50 on ImageNet dataset. We investigate $\gamma_c$ and $\beta_c$ at different depths of the network including layer1.0.bn1, layer2.0.bn1,layer3.0.bn1 and layer4.0.bn1. \textbf{(a-d)} shows BW enlarges $\beta_c$ when $\beta_c>0$ while reducing $\beta_c$ when $\beta_c\leq 0$.  \textbf{(e-h)} shows that BW consistently increases the magnitude of $\gamma_c$ across the network.}
\label{fig:BW-betagamma}
\vspace{-0.1in}
\end{figure*}

\begin{figure*}
\begin{center}
\includegraphics[scale=0.40]{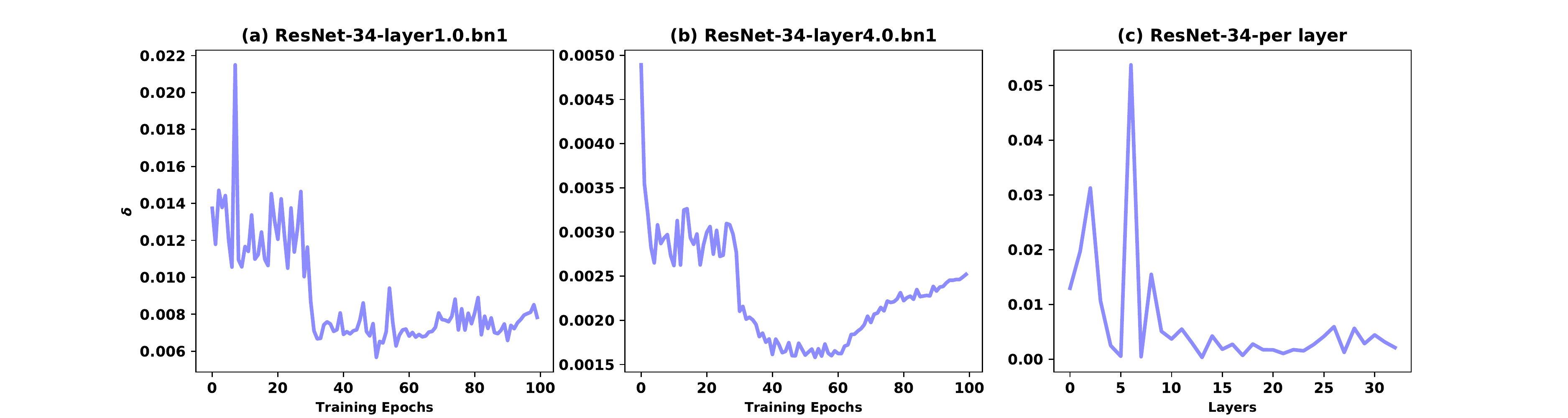}
\end{center}
\vspace{-0.1in}
  \caption{Experimental observation of the values of $\delta$ defined in proposition \ref{remark:3}. Results are obtained by tranining ResNet-34 on ImageNet dataset. (\textbf{a} \& \textbf{b}) investigate $\delta$ at different depths of the network including layer1.0.bn1 and layer4.0.bn1 respectively. (\textbf{c}) visualize $\delta$ for each layer of ResNet-34. We see that $\delta$ in proposition \ref{remark:3} is always a small positive constant.}
\label{fig:delta}
\vspace{-0.1in}
\end{figure*}

For (1), through Eqn.(\ref{eq:gamma2}), we acquire $|\hat{\gamma}_c| =\frac{1}{2} |3-\sum_{j=1}^C\frac{\gamma_j^2\rho_{cj}}{\left\|\boldsymbol{\gamma}\right\|_2^2}||\gamma_c|$. Therefore,  $|\hat{\gamma}_c|\rightarrow 0$ if $|{\gamma}_c|\rightarrow 0$. On the other hand, by Eqn.(\ref{eq:beta2}), we have $ \hat{\beta}_c \approx \frac{1}{2}(3-\frac{\gamma_c^2}{\left\|\boldsymbol{\gamma}\right\|_2^2})\beta_c < \beta_c \leq 0$. Here  we assume that $\rho_{cd}=1$ if $c=d$ and $0$ otherwise. Note that the assumption is plausible by Fig.\ref{fig:analysis-6} (e \& f) in main text from which we see that the correlation among channel features will gradually decrease during training. We also empirically verify these two conclusions by Fig.\ref{fig:BW-betagamma}. From Fig.\ref{fig:BW-betagamma} we can see that $|\hat{\gamma}_c| \geq |\gamma_c|$ where the equality holds iff  $|\gamma_c|=0$, and $\hat{\beta}_c$ is larger than $\beta_c$ if $\beta_c$ is positive, and vice versa. By Proposition \ref{remark:1}, we arrive at
\begin{equation}\label{eq:probcase1}
    P(\hat{X}_c>\delta) < P(\hat{X}_c>0) =0
\end{equation}
where the first `$>$' holds since $\delta$ is a small positive constant and `$=$' follows from $|\hat{\gamma}_c|\rightarrow 0$ and  $\hat{\beta}_c \leq 0$.
For (2), to show $P({\hat{{X}}}_c > \delta) > P({\tilde{{X}}}_c > \delta)$, we only need to prove $P({\bar{{X}}}_c > \frac{\delta-\hat{\beta}_c}{|\hat{\gamma}_c|}) > P({\bar{{X}}}_c > \frac{\delta-{\beta}_c}{|{\gamma}_c|})$, which is equivalent to $\frac{\delta-\hat{\beta}_c}{|\hat{\gamma}_c|} < \frac{\delta-{\beta}_c}{|{\gamma}_c|}$. To this end, we calculate
\begin{equation}\label{eq:probcase2}
\begin{split}
    \frac{|{\gamma}_c|\hat{\beta}_c-|\hat{\gamma}_c|{\beta}_c}{|\hat{\gamma}_c| - |{\gamma}_c|}&=
    \frac{|\gamma_c|\frac{1}{2}(3\beta_c-(\sum_{j=1}^C\frac{\gamma_j\beta_j\rho_{cj}}{\left\|\boldsymbol{\gamma}\right\|_2^2})\gamma_c) - \frac{1}{2} (3-\sum_{j=1}^C\frac{\gamma_j^2\rho_{cj}}{\left\|\boldsymbol{\gamma}\right\|_2^2})|\gamma_c|\beta_c}{\frac{1}{2} (3-\sum_{j=1}^C\frac{\gamma_j^2\rho_{cj}}{\left\|\boldsymbol{\gamma}\right\|_2^2})|\gamma_c| - |\gamma_c|}\\
    &= \frac{\sum_{j=1}^C\frac{\gamma_j\beta_j\gamma_c\rho_{cj}}{\left\|\boldsymbol{\gamma}\right\|_2^2} - \sum_{j=1}^C\frac{\gamma_j^2\beta_c\rho_{cj}}{\left\|\boldsymbol{\gamma}\right\|_2^2}}{1-\sum_{j=1}^C\frac{\gamma_j^2\rho_{cj}}{\left\|\boldsymbol{\gamma}\right\|_2^2}} \\
    &= \frac{\sum_{j=1}^C\frac{\gamma_j(\beta_j\gamma_c-\gamma_j\beta_c)\rho_{cj}}{\left\|\boldsymbol{\gamma}\right\|_2^2}}{1-\sum_{j=1}^C\frac{\gamma_j^2\rho_{cj}}{\left\|\boldsymbol{\gamma}\right\|_2^2}} \leq \frac{\frac{1}{\left\|\boldsymbol{\gamma}\right\|_2}\sqrt{\sum_{j=1}^C{(\beta_j\gamma_c-\gamma_j\beta_c)^2\rho^2_{cj}}}}{1-\sum_{j=1}^C\frac{\gamma_j^2\rho_{cj}}{\left\|\boldsymbol{\gamma}\right\|_2^2}} \\
    &= \frac{{\left\|\boldsymbol{\gamma}\right\|_2}\sqrt{\sum_{j=1}^C{(\beta_j\gamma_c-\gamma_j\beta_c)^2\rho^2_{cj}}}}{\left\|\boldsymbol{\gamma}\right\|_2^2-\sum_{j=1}^C{\gamma_j^2\rho_{cj}}} = \delta
\end{split}
\end{equation}
where the `$\leq$' holds due to the Cauchy–Schwarz inequality. By Eqn.(\ref{eq:probcase2}), we derive that $|{\gamma}_c|(\delta-\hat{\beta}_c) \leq |\hat{\gamma}_c|(\delta-{\beta}_c)$ which is exactly what we want.
Lastly, we empirically verify that $\delta$ defined in Proposition \ref{remark:3} is a small positive constant. In fact, $\delta$ represents the minimal activation feature value (\ie $\hat{X}_c=\hat{\gamma}_c\bar{X}_c+\hat{\beta}_c\geq \delta$ by definition). We visualize the value of $\delta$ in shallow and deep layers in ResNet-34 during the whole training stage and value of $\delta$ of each layer in trained ResNet-34 on ImageNet dataset in Fig.\ref{fig:delta}. As we can see, $\delta$ is always a small positive number in the training process. We thus empirically set $\delta$ as $0.05$ in all experiments.

To conclude, by Eqn.(\ref{eq:probcase1}), BWCP can keep the activation probability of unimportant channel unchanged; by Eqn.(\ref{eq:probcase2}), BWCP can increase the activation probability of important channel. In this way, the proposed BWCP can pursuit a compact deep model with good performance.


%-------------------------------------------------------------------------
\subsection{Solution to Residual Issue}\label{sec:A.5}
\begin{figure}
\begin{center}
\includegraphics[scale=0.8]{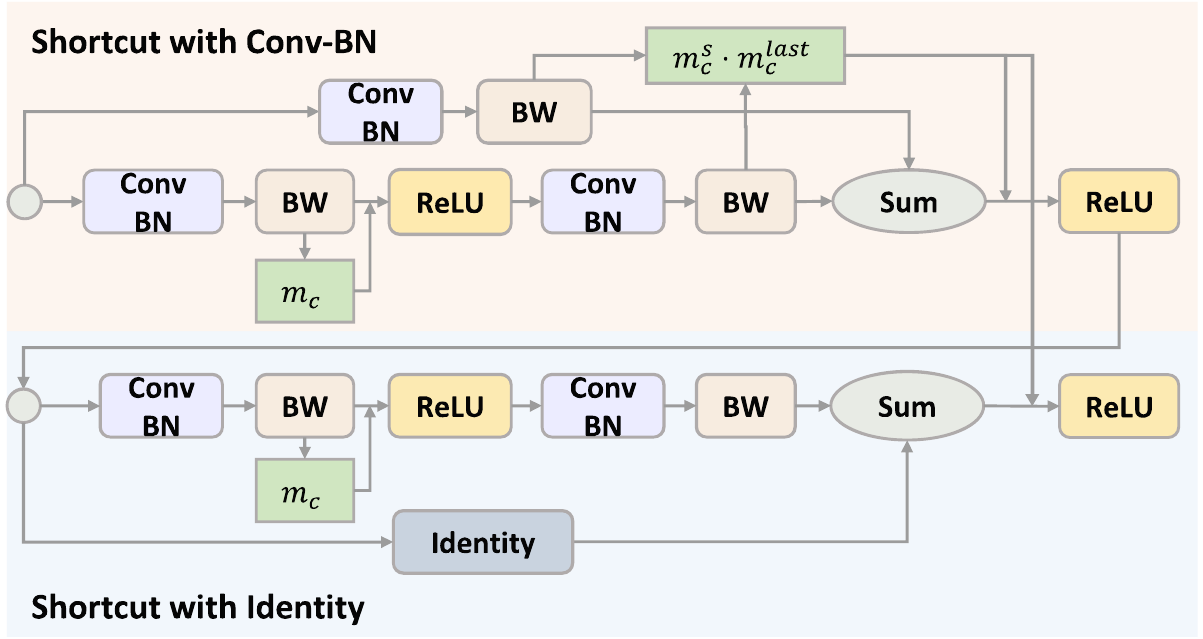}
\end{center}
\vspace{-0.1in}
  \caption{Illustration of BWCP with shortcut in basic block structure of ResNet. For shortcut with Conv-BN modules, we use a simple strategy that lets BW layer in the last convolution layer and shortcut share the same mask. For shortcut with identity mappings, we use the mask in previous layer.}
\label{fig:BWCP-shortcut}
\vspace{-0.1in}
\end{figure}
The recent advanced CNN architectures usually have residual blocks with shortcut connections \cite{he2016deep,huang2017densely}. As shown in Fig.\ref{fig:BWCP-shortcut}, the number of channels in the last convolution layer must be the same as in previous blocks due to the element-wise summation. Basically, there are two types of residual connections, \ie shortcut with downsampling layer consisting of Conv-BN modules, and shortcut with identity. For shortcut with Conv-BN modules, the proposed BW technique is utilized in downsampling layer to generate pruning mask $m_c^s$. Furthermore, we use a simple strategy that lets BW layer in the last convolution layer and shortcut share the same mask as given by $m_c=m_c^s\cdot m_c^{last}$ where ${m}_c^{last}$ and ${m}_c^s$ denote masks of the last convolution layer. For shortcut with identity mappings, we use the mask in the previous layer.  In doing so, their activated output channels must be the same.

\begin{algorithm}[tb]
	\caption{Forward Propagation of the proposed BWCP.}
	\label{alg_forward}
	\begin{algorithmic}[1]
		\begin{small}
			\STATE \textbf{Input}: mini-batch inputs $ \mathbf{x} \in \mathbb{R}^{N \times C \times H \times W} $.
			\STATE \textbf{Hyperparameters}: momentum $g$ for calculating root inverse of covariance matrix, iteration number $T$.
			\STATE \textbf{Output}: the activations $ {\mathbf{x}}^{out} $ obtained by BWCP.
			\STATE	calculate standardized activation: $\{\bar{\mathbf{x}}_c\}_{c=1}^C$ in Eqn.(\ref{eq:conv-norm-relu}).
			\STATE	calculate the output of BN layer: $\tilde{\mathbf{x}}_c = \gamma_c\bar{\mathbf{x}}_c+\beta_c$. 
			\STATE	calculate normalized covariance matrix: $\mathbf{\Sigma}_N= =\frac{\boldsymbol{\gamma}\boldsymbol{\gamma}\tran}{\left\|\boldsymbol{\gamma}\right\|_2^2}\odot\frac{1}{NHW}\sum_{n,i,j=1}^{N,H,W} \bar{\mathbf{x}}_{nij} \bar{\mathbf{x}}_{nij}\tran$
			\STATE $\mathbf{\Sigma}_0=\mathbf{I}$.
			\FOR {$k = 1 ~~to ~~T $}
			\STATE $\mathbf{\Sigma}_{k}=\frac{1}{2} (3 \mathbf{\Sigma}_{k-1} - \mathbf{\Sigma}_{k-1}^{3} \Sigma_{N})$
			\ENDFOR	
			\STATE  calculate whitening matrix for training: $\mathbf{\Sigma}_N^{-\frac{1}{2}} = \mathbf{\Sigma}_T$.
			\STATE  calculate whitening matrix for inference: $\hat{\mathbf{\Sigma}}_N^{-\frac{1}{2}} \leftarrow (1-g)\hat{\mathbf{\Sigma}}_N^{-\frac{1}{2}} + g \mathbf{\Sigma}_N^{-\frac{1}{2}}$.
			\STATE  calculate  whitened output: $\hat{\mathbf{x}}_{nij} = \mathbf{\Sigma}_N^{-\frac{1}{2}} \tilde{\mathbf{x}}_{nij}$.
		\STATE	calculate equivalent scale and bias defined by BW: $\hat{\boldsymbol{\gamma}} = \Sigmainv\boldsymbol{\gamma}$ and $\hat{\boldsymbol{\beta}} = \Sigmainv\boldsymbol{\beta}$.
			\STATE	calculate the activation probability by Proposition \ref{remark:2} with $\hat{\boldsymbol{\gamma}}$ and $\hat{\boldsymbol{\beta}}$, obtain soft masks $\{m_c\}_{c=1}^C$ by Eqn.(\ref{eq:soft-gate}).
			\STATE	calculate the output of BWCP: ${\mathbf{x}}_{c}^{\mathrm{out}} = {\hat{\mathbf{x}}_{c}} \odot {{m}_c}$. 

		\end{small}
	\end{algorithmic}
\end{algorithm}

%-------------------------------------------------------------------------

%-------------------------------------------------------------------------
\subsection{Back-propagation of BWCP}\label{sec:A.6}
\begin{figure}
\begin{center}
\includegraphics[scale=0.6]{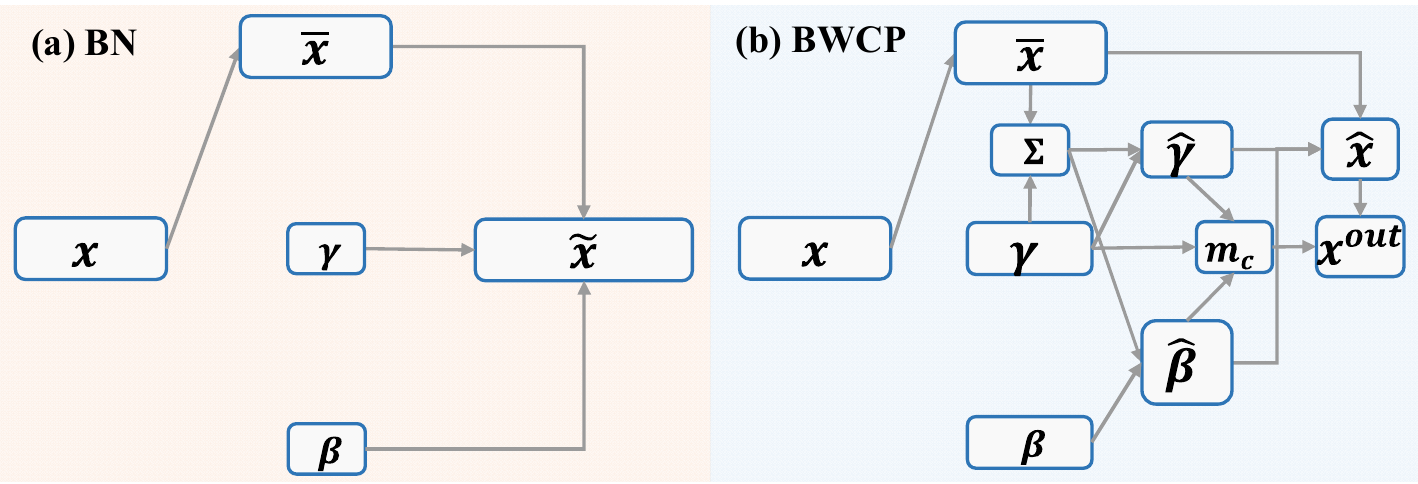}
\end{center}
\vspace{-0.1in}
  \caption{Illustration of forward propagation of \textbf{(a)} BN and \textbf{(b)} BWCP. The proposed BWCP prunes CNNs by replacing original BN layer with BWCP module.}
\label{fig:BWCP-back}
\vspace{-0.1in}
\end{figure}

The forward propagation of BWCP can be represented by Eqn.(3-4) and Eqn.(9) in the main text (see detail in Table \ref{alg_forward}), all of which  define differentiable transformations. Here we provide the back-propagation of BWCP. By comparing the forward representation of BN and BWCP in Fig.\ref{fig:BWCP-back}, we need to back-propagate the gradient $\frac{\partial \LL}{\partial {\mathbf{x}}^{out}_{nij}}$ to $\frac{\partial \LL}{\partial \bar{\mathbf{x}}_{nij}}$
%, \frac{\partial \LL}{\partial {\mathbf{\gamma}}}$ and $\frac{\partial \LL}{\partial {\mathbf{\beta}}}$ 
for backward propagation of BWCP. For simplicity, we neglect the subscript `$nij$'.
%Hence, we need to calculate $\frac{\partial \hat{\mathbf{x}}_{nij}}{\partial \bar{\mathbf{x}}_{nij}}, \frac{\partial \hat{\mathbf{x}}_{nij}}{\partial {\mathbf{\gamma}}}$ and $\frac{\partial \hat{\mathbf{x}}_{nij}}{\partial {\mathbf{\beta}}}$. 
%
\begin{table*}[!t]
	\centering
	\caption{
		Performance of our BWCP on different base models compared with other approaches on CIFAR-100 dataset. 
	}%Slimming and Variational Pruning
	%\vspace{-0.1in}
	\scalebox{0.70}{
		\begin{tabular}{c c c c c c c c }		
			\toprule
			Model & Mothod  & Baseline Acc. (\%) & Acc. (\%) & Acc. Drop & Channels $\downarrow$ (\%) & Model Size $\downarrow$ (\%) & FLOPs $\downarrow$ (\%) \\
			\hline %\hline
			\multirow{3}{*}{ResNet-164}& Slimming*~\cite{liu2017learning}& 77.24 & 74.52 & 2.72 &\textbf{60} & \textbf{29.26} & \textbf{47.92} \\
			%\cdashline{2-9}
			 & SCP~\cite{kang2020operation}   & 77.24 & 76.62 & 0.62 & 57 & 28.89 & 45.36 \\
			 & BWCP (Ours)   & 77.24 & 76.77 & \textbf{0.47} & 41 & 21.58 & 39.84 \\
			\hline %\hline
			\multirow{4}{*}{DenseNet-40}& Slimming*~\cite{liu2017learning}& 74.24 & 73.53 & 0.71 & \textbf{60} & 54.99 & \textbf{50.32} \\
		%\cdashline{2-9}
			 &Variational Pruning~\cite{zhao2019variational}  & 74.64 & 72.19 & 2.45 & 37 & 37.73 & 22.67 \\
			  & SCP~\cite{kang2020operation}   & 74.24 & 73.84 & 0.40 & \textbf{60} & \textbf{55.22} & 46.25 \\ 
		  & 	BWCP (Ours)   & 74.24 & 74.18 & \textbf{0.06} & 54 & 53.53 & 40.40 \\
			\hline %\hline
			\multirow{2}{*}{VGGNet-19}& Slimming* ~\cite{liu2017learning}& 72.56 & 73.01 & -0.45 &  \textbf{50} & \textbf{76.47} & \textbf{38.23} \\
			%\cdashline{2-9}
			%SCP~\cite{kang2020operation} & VGGNet-19  & X & 72.56 & 72.99 & -0.43 & \textbf{51} & \textbf{77.52} & \textbf{40.92} \\
			 & BWCP (Ours)  & 72.56 & 73.20 & \textbf{-0.64} & 23 & 41.00 & 22.09 \\
			\hline %\hline
			\multirow{4}{*}{VGGNet-16}  & Slimming* ~\cite{liu2017learning}& 73.51 & 73.45 & 0.06 & \textbf{40} & \textbf{66.30} & 27.86 \\
			%\cdashline{2-9}
		 &	Variational Pruning~\cite{zhao2019variational}  & 73.26 & 73.33 & -0.07 & 32 &37.87 & 18.05 \\
			%SCP~\cite{kang2020operation} & VGGNet-16  & X & 73.51 & 73.86 & -0.35 & \textbf{52} & \textbf{80.14} & \textbf{51.45} \\
			 &BWCP (Ours)   & 73.51 & 73.60 & \textbf{-0.09} & 34 & 58.16 & \textbf{34.46} \\
			\bottomrule
		\end{tabular}
	}
	\label{table:cifar100_table_1}
\end{table*}

By chain rules,  we have
\begin{equation}\label{eq:grad-barx}
    \frac{\partial \LL}{\partial \bar{\mathbf{x}}} = \hat{\boldsymbol{\gamma}}\odot\mathbf{m}\odot \frac{\partial \LL}{\partial {\mathbf{x}}^{out}} + \frac{\partial \LL}{\partial \bar{\mathbf{x}}}(\SigmainvN)
\end{equation}
where $\frac{\partial \LL}{\partial \bar{\mathbf{x}}}(\SigmainvN)$ denotes the gradient \textit{w.r.t.} $ \bar{\mathbf{x}}$ back-propagated through $\SigmainvN$. To calculate it, we first obtain the gradient \textit{w.r.t.} $\SigmainvN$ as given by
\begin{equation}\label{eq:grad-sigmainv}
    \frac{\partial \LL}{\partial \SigmainvN} = \boldsymbol{\gamma}\frac{\partial \LL}{\partial {\hat{\boldsymbol{\gamma}}}}\tranT + \boldsymbol{\beta}\frac{\partial \LL}{\partial {\hat{\boldsymbol{\beta}}}}\tranT
\end{equation}
where
\begin{equation}
    \frac{\partial \LL}{\partial {\hat{\boldsymbol{\gamma}}}} = \bar{\mathbf{x}}\odot\mathbf{m}\odot \frac{\partial \LL}{\partial {\mathbf{x}}^{out}} +  
    \frac{\partial {\mathbf{m}}}{\partial \hat{\boldsymbol{\gamma}}}(\hat{\mathbf{x}}\odot \frac{\partial \LL}{\partial {\mathbf{x}}^{out}})
\end{equation}
and
\begin{equation}
    \frac{\partial \LL}{\partial {\hat{\boldsymbol{\beta}}}} = \mathbf{m} +  
    \frac{\partial {\mathbf{m}}}{\partial \hat{\boldsymbol{\beta}}}(\hat{\mathbf{x}}\odot \frac{\partial \LL}{\partial {\mathbf{x}}^{out}})
\end{equation}
The remaining thing is to calculate $\frac{\partial {\mathbf{m}}}{\partial \hat{\boldsymbol{\gamma}}}$ and $\frac{\partial {\mathbf{m}}}{\partial \hat{\boldsymbol{\beta}}}$. Based on the Gumbel-Softmax transformation, we arrive at
\begin{align}
\frac{\partial {{m}_c}}{\partial \hat{{\gamma}}_d}&=
    \left\{
\begin{array}{l}
\frac{-m_c(1-m_c)f(\hat{\gamma}_c, \hat{\beta}_c)}{\tau P(\hat{X}_C>0)(1-P(\hat{X}_C>0))}\frac{\beta_c\gamma_c}{|\gamma_c|^2}\, \mathrm{if} \,d = c\\
0, \, \mathrm{otherwise}
\end{array}
\right.  \\
\frac{\partial {{m}_c}}{\partial \hat{{\beta}_d}}&=
    \left\{
\begin{array}{l}
\frac{m_c(1-m_c)f(\hat{\gamma}_c, \hat{\beta}_c)}{\tau P(\hat{X}_C>0)(1-P(\hat{X}_C>0))},\, \mathrm{if}\, d=c\\
0, \, \mathrm{otherwise} 
\end{array}
\right.
\end{align}
where $f(\hat{\gamma}_c, \hat{\beta}_c)$ is the probability density function of R.V. $\hat{X}_c$ as written in Eqn.(2) of main text. 

To proceed, we deliver the gradient \textit{w.r.t.} $\SigmainvN$ in Eqn.(\ref{eq:grad-sigmainv}) to ${\mathbf{\Sigma}}$ by Newton Iteration in Eqn.(6) of main text. Note that $\SigmainvN = \Sigma_T$, we have
\begin{align}\label{eq:grad-sigma}
     \frac{\partial \LL}{\partial \mathbf{\Sigma}_N}=
     -\frac{1}{2}\sum_{k=1}^T(\mathbf{\Sigma}_{k-1}^3)\tranT  \frac{\partial \LL}{\partial \mathbf{\Sigma}_k}
\end{align}
where $\frac{\partial \LL}{\partial \mathbf{\Sigma}_k}$ can be calculated by following iterations:
\begin{align*}
    \frac{\partial \LL}{\partial \mathbf{\Sigma}_{k-1}}&=\frac{3}{2}\frac{\partial \LL}{\partial \mathbf{\Sigma}_k}-\frac{1}{2}\frac{\partial \LL}{\partial \mathbf{\Sigma}_k}(\mathbf{\Sigma}_{k-1}^2\mathbf{\Sigma})\tranT
    -\frac{1}{2}(\mathbf{\Sigma}_{k-1}^2)\tranT\frac{\partial \LL}{\partial \mathbf{\Sigma}_k}\mathbf{\Sigma}\tranT \\
    &-\frac{1}{2}(\mathbf{\Sigma}_{k-1})\tranT\frac{\partial \LL}{\partial \mathbf{\Sigma}_k}(\mathbf{\Sigma}_{k-1}\mathbf{\Sigma})\tranT\, k=T,\cdots, 1.
\end{align*}
Given the gradient \textit{w.r.t.} $\mathbf{\Sigma}$ in Eqn.(\ref{eq:grad-sigma}), we can calculate the gradient \textit{w.r.t.} $ \bar{\mathbf{x}}$ back-propagated through $\SigmainvN$ in Eqn.(\ref{eq:grad-barx}) as follows
\begin{equation}\label{eq:grad-last}
    \frac{\partial \LL}{\partial \bar{\mathbf{x}}}(\SigmainvN)=
    (\frac{\boldsymbol{\gamma}\boldsymbol{\gamma}\tran}{\left\|\boldsymbol{\gamma}\right\|^2} \odot (\frac{\partial \LL}{\partial \mathbf{\Sigma}_N} +\frac{\partial \LL}{\partial \mathbf{\Sigma}_N}\tranT)) \bar{\mathbf{x}}
\end{equation}
Based on Eqn.(\ref{eq:grad-barx}-\ref{eq:grad-last}), we obtain the back-propagation of BWCP.
\section{More Details about Experiment}

\subsection{Training Configuration}\label{sec:B.1}
\textbf{Training Setting on ImageNet}.
All networks are trained using 8 GPUs with a mini-batch of $32$ per GPU. We train all the architectures from scratch for $100$ epochs using stochastic gradient descent (SGD) with momentum $0.9$ and weight decay 1e-4. The base learning rate is set to $0.1$ and is multiplied by $0.1$ after $30,60$ and $90$ epochs. The fine-tuning procedure uses the same configuration except that the initial learning rate is set to $0.01$. The coefficient of sparse regularization $\lambda_1$ and $\lambda_2$ are set to 7e-5 and 3.5e-5. Besides, the covariance matrix in the proposed BW technique is calculated within each GPU. Like \cite{huang2019iterative}, we also use group-wise decorrelation with group size $16$ across the network to improve the efficiency of BW. 

\textbf{Training setting on CIFAR-10 and CIFAR-100}.  We train all models on CIFAR-10 and CIFAR-100 with a batch size of $64$ on a single GPU for $160$ epochs with momentum $0.9$ and weight decay 1e-4. The initial learning rate is 0.1 and is decreased by $10$ times at $80$ and $120$ epoch. The coefficient of sparse regularization $\lambda_1$ and $\lambda_2$ are set to 4e-5 and 8e-5 for CIFAR-10 dataset and 7e-6 and 1.4e-5 for CIFAR-100 dataset.

\subsection{More Results of BWCP}\label{sec:B.2}
The results of BWCP on CIFAR-100 dataset is reported in Table \ref{table:cifar100_table_1}. As we can see, our approach BWCP achieves the lowest accuracy drops and comparable FLOPs reduction compared with existing channel pruning methods in all tested base models. 

%{\small
%\bibliography{MSegbib}
%}

%%%%%%%%%%%%%%%%%%%%%%%%%%%%%%%%%%%%%%%%%%%%%%%%%%%%%%%%%%%%%%%%%%%%%%%%%%%%%%%
%%%%%%%%%%%%%%%%%%%%%%%%%%%%%%%%%%%%%%%%%%%%%%%%%%%%%%%%%%%%%%%%%%%%%%%%%%%%%%%
\end{document}